\documentclass{article}

\usepackage{arxiv_preprint,times}

\usepackage{amsmath,amsfonts,bm}

\def\eqref#1{equation~\ref{#1}}

\def\1{\bm{1}}

\DeclareMathAlphabet{\mathsfit}{\encodingdefault}{\sfdefault}{m}{sl}
\SetMathAlphabet{\mathsfit}{bold}{\encodingdefault}{\sfdefault}{bx}{n}

\usepackage{amsmath,amssymb,amsthm}
\usepackage{booktabs}
\usepackage{graphicx}
\usepackage{wrapfig}
\usepackage{needspace}
\usepackage{float}
\usepackage{placeins}
\usepackage{capt-of}
\usepackage{microtype}
\usepackage{xcolor}
\usepackage{hyperref}
\renewcommand{\eqref}[1]{\textup{(\ref{#1})}}
\usepackage{url}
\hypersetup{colorlinks=true,linkcolor=black,citecolor=black,urlcolor=blue}

\newtheorem{theorem}{Theorem}
\newtheorem{lemma}[theorem]{Lemma}
\newtheorem{corollary}[theorem]{Corollary}
\newtheorem{proposition}[theorem]{Proposition}
\theoremstyle{definition}

\newcommand{\Rnu}{R_{\nu}}
\newcommand{\Atwo}{A_{R,\nu}}
\newcommand{\Ftwo}{F_{R,\nu}}
\newcommand{\phinu}{\Phi_{\nu}}
\newcommand{\norm}[1]{\left\lVert #1\right\rVert}
\newcommand{\TV}{\operatorname{TV}}

\title{SAME LOSS, DIFFERENT GRADIENTS}
\author{\parbox{\dimexpr\textwidth-2\tabcolsep\relax}{\centering
Ningkang Peng\textsuperscript{1}, Xiaoqian Peng\textsuperscript{2}, Yifan He\textsuperscript{1}, Anjie Hu\textsuperscript{1},\\
Chao Tan\textsuperscript{1}, Peirong Ma\textsuperscript{1}, Yanhui Gu\textsuperscript{1}\\[0.5em]
\normalfont\textsuperscript{1}Nanjing Normal University\\
\textsuperscript{2}Nanjing University of Chinese Medicine\\
\texttt{nkpeng@nnu.edu.cn}, \texttt{gu@njnu.edu.cn}}}
\iclrfinalcopy

\begin{document}
\maketitle

\begin{abstract}
Differentiable learning typically assumes that the scalar objective evaluated in the forward pass and the gradient supplied to the optimizer in the backward pass describe the same mathematical object. We show that this correspondence can fail when probabilistic objectives rely on finite special-function recurrences, custom backward rules, and numerical clipping. In high-dimensional von Mises--Fisher learning, real numerical implementations can produce identical forward scores and losses at the same learning state while supplying different gradients and following different optimization trajectories. We characterize the structure of this mismatch in finite-start Bessel recurrence and show that classwise radial mismatch can compose through probabilities into a locally nonconservative update field. Evaluating the accuracy of special-function values and derivatives separately is therefore insufficient to characterize the realized learning objective. Motivated by this observation, we introduce AR/FR, a fixed-depth analytic realization that constructs a potential and its derivative jointly, ensuring forward--backward coherence by construction. We establish a uniform cubic-order error bound relative to the exact Bessel ratio over the entire nonnegative concentration axis and propagate this guarantee to learning scores and objectives. As representation dimension increases, the original finite recurrence becomes sequentially deeper, whereas the worst-case AR/FR error guarantee tightens cubically, jointly providing coherence, certified fidelity, and fixed-depth computation. These results suggest that a differentiable numerical primitive is defined by both the values it realizes and the derivatives it actually supplies to the optimizer; together, they constitute the numerical realization of the learning algorithm.
\end{abstract}

\section{Introduction}

In differentiable machine learning, a scalar objective serves two roles: the forward pass evaluates its value, and the backward pass supplies its derivative to the optimizer to determine the next update. We usually assume that these processes describe the same mathematical object. If the program realizes a numerical objective $\widetilde L$, its backward should supply $\nabla\widetilde L$. In ordinary automatic differentiation, this correspondence is typically guaranteed by the computation graph \citep{baydin2018automatic}. When the objective contains high-order special functions, finite recurrences, clipping, or a custom backward, however, that guarantee can fail. Two implementations may then return identical forward scores and losses while supplying different gradients and ultimately following different training trajectories. \textbf{The same forward loss need not imply the same learning process.}

This issue is particularly relevant to probabilistic learning, where objectives often depend on normalization constants and special functions that are difficult to evaluate efficiently. High-dimensional von Mises--Fisher (vMF) learning is a representative example \citep{Scott_2021_ICCV,sra2012note}: its scalar potential contains a high-order modified Bessel function, and its analytic derivative is the adjacent-order Bessel ratio
\[
 R_\nu(x)=\frac{I_{\nu+1}(x)}{I_\nu(x)}.
\]
The Bessel order grows with representation dimension, motivating finite recurrences, ratio approximations, and custom backward rules in practical learning systems \citep{amos1974computation,du2024proco,he2025patt}. This raises a concrete question of numerical coherence: after approximation, do the forward potential and supplied derivative retain their original derivative relation?

We observe this decoupling in real vMF learning implementations. In PATT \citep{he2025patt}, the original implementation and automatic differentiation of the same finite forward produce elementwise-identical scores and losses from identical model states and inputs, yet supply different feature gradients. As the branches continue to update independently, their parameter and representation trajectories separate. An additional audit of ProCo \citep{du2024proco} detects the same type of mismatch in a second implementation, with a magnitude that depends on the numerical states visited during learning. \textbf{These observations move the question beyond local approximation error to the numerical realization of the learning objective: does the value reported by the program remain coherent with the derivative received by the optimizer?}

To explain this incoherence, we analyze the internal structure of finite-start Bessel recurrence. We prove that the sign of the ratio error is determined exactly by the parity relation between recurrence depth and target order, and characterize its high-concentration asymptotics. Clipping bounds the supplied ratio but does not restore its derivative relation with the forward potential. Consequently, deeper recurrence can improve fidelity to the exact Bessel ratio without monotonically improving coherence. After Softmax composition, forward probabilities couple with mismatched classwise radial rules, potentially producing an asymmetric Jacobian and a locally nonconservative update field. This establishes a mechanism linking special-function mismatch to the geometry of learning updates.

The mismatch also accompanies a dimension-dependent computational structure. Sequential recurrence depth grows with Bessel order, while high orders are precisely where the analytic approximation becomes tighter. Thus, as the original recurrence becomes deeper in high-dimensional vMF learning, a fixed-depth analytic replacement gains increasingly favorable approximation conditions. This motivates a paired analytic realization that avoids recurrence over the order.

We describe such a realization through two complementary requirements. \emph{Coherence} asks whether the supplied backward differentiates the currently realized scalar objective. \emph{Fidelity} asks whether this numerical surrogate remains sufficiently close to the intended analytic vMF objective. These properties are independent: an accurate forward approximation can be paired with a backward rule that does not belong to it, while an internally coherent numerical object can approximate the analytic target poorly. A reliable differentiable numerical primitive must address both questions.

We introduce \textbf{AR/FR}, which directly constructs a potential--derivative pair instead of designing the forward potential and backward ratio independently, so that
\[
 F_R'(x)=A_R(x)
\]
holds in exact arithmetic. Coherence is therefore a property of the construction itself. We further prove a uniform cubic-order fidelity bound for AR relative to the exact Bessel ratio over the entire nonnegative concentration axis. Integration transfers this guarantee to the potential endpoint differences used in vMF scores and then to the fixed-state learning objective. AR/FR also has fixed computational depth, eliminating sequential recurrence over Bessel order. Since feature dimension is linearly related to the order, increasing dimension lengthens the original recurrence chain while tightening the worst-case AR/FR error guarantee as $O(p^{-3})$. \textbf{The high-dimensional regime in which recurrence becomes deeper is also the regime in which the analytic replacement receives a stronger theoretical guarantee.}

Real learning states, high-precision numerical comparisons, and complete training experiments validate this mechanism and analytic realization. They show that a paired realization can eliminate forward--backward incoherence while maintaining competitive task performance.

Our main contributions are:
\begin{enumerate}
\item \textbf{Numerical incoherence and its mechanism.} We identify identical forward values with different gradients in real vMF learning implementations, characterize finite-start parity, high-concentration behavior, and clipping, and show that classwise mismatch can compose through probabilities into a locally nonconservative update field.
\item \textbf{A coherent and certified paired vMF realization.} AR/FR guarantees forward--backward coherence through $F_R'=A_R$ and provides a uniform cubic-order fidelity bound over the nonnegative concentration axis, together with its propagation to learning scores and objectives.
\item \textbf{Fixed-depth high-dimensional computation.} AR/FR replaces order-dependent recurrence with fixed-depth analytic computation, revealing a complexity--accuracy reversal as dimension increases: recurrence becomes deeper while the approximation guarantee tightens. This realization substantially reduces the cost of the practical vMF computation path.
\end{enumerate}

\begin{figure}[t]
\centering
\includegraphics[width=\linewidth]{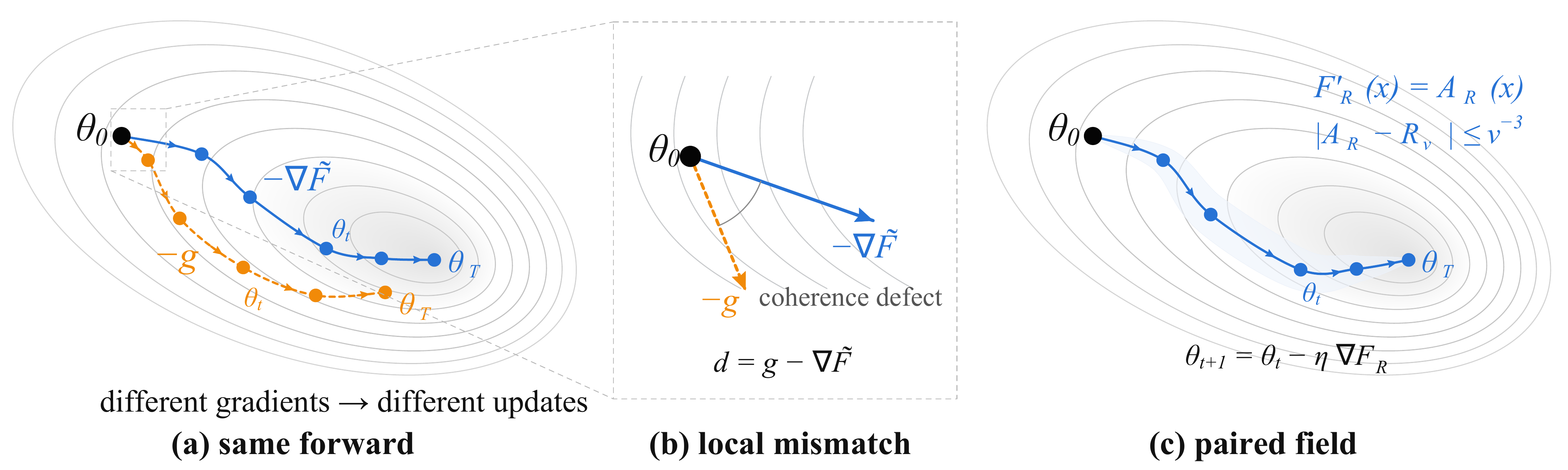}
\caption{\textbf{Same loss, different gradients: diagnosis and paired realization.}
(a) A common forward objective can supply different updates when backward rules differ.
(b) The coherence defect compares the supplied backward with the derivative of that forward.
(c) AR/FR constructs the potential and derivative together, combining coherence with a uniform fidelity bound.}
\label{fig:overview}
\end{figure}

\section{Coherence and fidelity of numerical realizations}
\label{sec:realization}
\subsection{Coherence, fidelity, and realized objectives}
Let $F$ denote an analytic scalar objective, $\widetilde F$ the forward function evaluated by a numerical program, and $g$ the supplied backward rule. Its \emph{coherence defect} is
\begin{equation}
 d(x)=g(x)-\widetilde F'(x).
 \label{eq:coherence-defect}
\end{equation}
The realization is coherent when $d=0$: its supplied backward differentiates its realized forward. In contrast, \emph{fidelity} measures proximity to the intended target, through derivative errors $|g-F'|$ and forward potential-difference errors. These properties are logically independent: a realization can approximate the target accurately without being internally coherent, or be coherent while approximating the target poorly.

Two implementations can evaluate the same $\widetilde F$ at a shared state but supply different backward rules. They then have identical forward values and losses yet induce different local updates. Throughout this paper, ``same loss'' denotes equality at a shared state; after separate updates, trajectories and subsequent losses need not remain identical. For potentials, we compare endpoint differences $F(b)-F(a)$, since additive constants do not affect learning.

\subsection{The vMF value--derivative relation}
Let the feature dimension be $p$ and $\nu=p/2-1$. For $x>0$, the vMF potential and its derivative satisfy
\begin{equation}
 \Phi_\nu(x)=\log I_\nu(x)-\nu\log x,
 \qquad \Phi_\nu'(x)=R_\nu(x)=\frac{I_{\nu+1}(x)}{I_\nu(x)},
 \label{eq:main-vmf-pair}
\end{equation}
where $I_\nu$ is the modified Bessel function of the first kind and $\Phi_\nu$ is extended continuously to $x=0$.

Fix a class state under the original stop-gradient semantics. Class $j$ has unit direction $\mu_j$ and concentration $\kappa_j\ge0$; $f$ is a query feature and $\tau>0$ is the temperature. Its scalar radius and score are
\begin{equation}
 r_j(f)=\left\lVert\kappa_j\mu_j+\frac f\tau\right\rVert,
 \qquad q_j(f)=\Phi_\nu(r_j(f))-\Phi_\nu(\kappa_j).
 \label{eq:realized-vmf-score}
\end{equation}
The score requires a forward potential difference and its radial derivative. Finite numerical evaluation can split this analytic pair into distinct computational paths.

\section{When finite recurrence breaks coherence}
\label{sec:diagnosis}
\subsection{Three numerical objects from one recurrence}
For integer orders $M>\nu\ge1$ and $x>0$, finite Miller backward recurrence uses
\[
 b_M=1,\quad b_{M+1}=0,\qquad b_{i-1}=\frac{2i}{x}b_i+b_{i+1}.
\]
It induces a finite forward potential $\widetilde\Phi_{\nu,M}$ and an adjacent-order ratio. The supplied radial backward may additionally clip that ratio:
\[
 \widetilde R_{\nu,M}(x)=\frac{b_{\nu+1}}{b_\nu},
 \qquad g_{\nu,M}(x)=\min\{\widetilde R_{\nu,M}(x),1\}.
\]
Thus, the potential, raw ratio, and supplied backward are three numerical objects. Coherence requires $g_{\nu,M}=\widetilde\Phi_{\nu,M}'$, which neither finite recurrence nor clipping guarantees. Appendix~\ref{app:released-details} specifies the finite forward and its normalization.

\subsection{Finite-start parity and clipping}
Each adjacent-ratio recurrence step is a strictly decreasing map on the positive axis and reverses the ordering relative to the exact tail. Consequently, we obtain the following all-axis result.
\begin{theorem}[Finite-start parity]
For integer $M>\nu\ge1$,
\begin{equation}
 \operatorname{sign}\bigl(\widetilde R_{\nu,M}(x)-R_\nu(x)\bigr)
 =(-1)^{M-\nu+1},\qquad x>0.
 \label{eq:main-sign}
\end{equation}
\end{theorem}
This all-axis parity law reduces to $(-1)^{\nu+1}$ under $M=2\nu$ (proof in Appendix~\ref{app:miller-sign-proof}). It characterizes ratio fidelity; coherence additionally depends on the finite-forward derivative.

At high concentration with $M=2\nu$, the raw ratio grows linearly for odd $\nu$ and decays inversely for even $\nu$, while the exact ratio approaches one (Appendix~\ref{app:miller-parity-proof}). Clipping bounds the supplied value but does not restore the potential relation. Increasing depth can improve raw ratio fidelity without monotonically improving the clipped coherence defect, which also depends on the finite-forward derivative and clipping regime. Section~\ref{sec:mismatch-regimes} examines these conditions numerically.

\subsection{Composition can change update-field geometry}
\label{sec:ce-integrability}
A scalar radial rule $g(r)$ admits a one-dimensional primitive, so radial mismatch alone does not rule out an underlying objective. The obstruction appears after classwise composition. Let $q_j=F(r_j)-F(\kappa_j)+b_j$, with fixed offsets $b_j$, and let $p_j=\exp(q_j)/\sum_k\exp(q_k)$. Write $v_j=\nabla_fq_j$ and $w_j=g(r_j)\nabla_fr_j$. Cross-entropy supplies the feature field
\[
 G(f)=\sum_j(p_j-\mathbf1[j=y])w_j.
\]
On a smooth neighborhood with fixed class state and positive radii, its antisymmetric Jacobian is
\begin{equation}
 J_G-J_G^\top=\bar v\bar w^\top-\bar w\bar v^\top,
 \qquad \bar v=\sum_jp_jv_j,\quad \bar w=\sum_jp_jw_j.
 \label{eq:ce-curl-obstruction}
\end{equation}
Forward probabilities depend on $F$, while feature updates use $g$. Softmax couples these radial realizations, and the antisymmetric Jacobian determines whether the composed field admits a local $C^2$ scalar potential.

When $g=F'$, we have $w_j=v_j$ and the expression vanishes, as required for a $C^2$ scalar-loss gradient field.

A valid fixed-state vMF construction yields a nonzero antisymmetric Jacobian component under the supplied backward, whereas differentiating the same forward gives zero. The full construction is given in Appendix~\ref{app:curl-proof}, with additional numerical controls in Appendix~\ref{app:mechanism-interventions}. This provides a local existence result: classwise mismatch can compose into a nonconservative feature-update field.

\begin{figure}[t]
\centering
\includegraphics[width=0.70\linewidth]{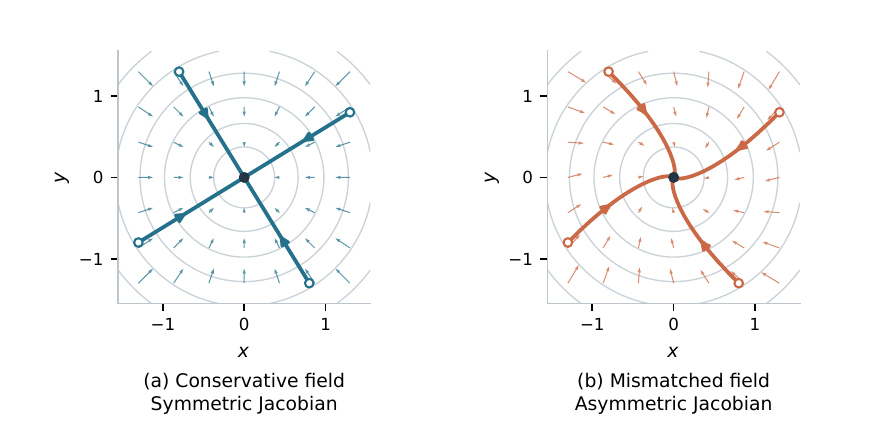}
\caption{\textbf{Local geometry of coherent and mismatched update fields.} Both panels share the same scalar forward surface. A matched backward yields a conservative gradient field with symmetric Jacobian, whereas a mismatched backward can introduce a local rotational component and asymmetric cross-derivatives. The figure is an analytic illustration.}
\label{fig:theoretical-field}
\end{figure}

\section{A paired analytic vMF realization}
\label{sec:f2}
\subsection{Paired analytic construction}
Analytic ratio approximations and their integrated forms have previously been used in vMF learning \citep{Scott_2021_ICCV}. Our goal is to make the numerical realization itself satisfy both forward--backward coherence and provable fidelity to the intended objective. We construct the potential and its derivative jointly. For $\nu>0$, $x\ge0$, and $s(x)=\sqrt{\nu^2+x^2}$, define
\begin{equation}
 A_{R,\nu}(x)=\frac{x}{\nu+s(x)}-\frac{x}{2s(x)^2}
 +\frac{x(4\nu^2-x^2)}{8s(x)^5}.
 \label{eq:a2}
\end{equation}
Let $F_{R,\nu}$ be its analytic primitive. The identity $F_{R,\nu}'=A_{R,\nu}$ makes the coherence defect vanish by construction. The forward potential and supplied derivative are therefore no longer approximated independently; both are generated from the same analytic object. The paired score is
\begin{equation}
 \widehat q_j(f)=F_{R,\nu}(r_j(f))-F_{R,\nu}(\kappa_j),
 \label{eq:f2-score}
\end{equation}
with the same fixed class state. Equation~\eqref{eq:f2} gives the closed form of $F_R$, and Appendix~\ref{app:system} describes stable endpoint evaluation.

\subsection{Uniform fidelity certificate}
Coherence alone does not establish fidelity to the analytic vMF target. The exact radial derivative $R_\nu$ satisfies a Riccati differential equation, which provides a direct way to analyze the error of the analytic approximation $A_{R,\nu}$.
\setcounter{theorem}{1}
\begin{theorem}[Uniform ratio error bound]
\label{thm:a2}
For every $\nu\ge10/9$ and $x\ge0$,
\begin{equation}
 |R_\nu(x)-A_{R,\nu}(x)|\le\nu^{-3}.
 \label{eq:a2-bound}
\end{equation}
\end{theorem}
The proof rescales $x=\nu z$ and substitutes $A_{R,\nu}$ into the Riccati equation for the exact Bessel ratio. The resulting residual is uniformly $O(\nu^{-3})$. Constant upper and lower barriers at $\pm\nu^{-3}$ are then shown to be inward-pointing for $\nu\ge10/9$, while the small-$x$ expansion initializes the error strictly between them. This prevents the approximation error from crossing either barrier and yields the bound over the entire nonnegative axis. The complete argument is given in Appendix~\ref{app:a2-proof}.

Unlike a grid-based numerical validation, this guarantee is independent of the concentration regime visited during training. It therefore complements the state-dependent coherence analysis in Section~\ref{sec:mismatch-regimes}: coherence defects can vary strongly across visited states, whereas the AR/FR fidelity certificate holds uniformly for every $x\ge0$.

\subsection{Learning-level fidelity and fixed-depth computation}
\label{sec:high-dimension}
Integrating the ratio certificate gives, for any $r,\kappa\ge0$,
\begin{equation}
 \left|[\Phi_\nu(r)-\Phi_\nu(\kappa)]-[F_{R,\nu}(r)-F_{R,\nu}(\kappa)]\right|
 \le\nu^{-3}|r-\kappa|.
 \label{cor:endpoint-fidelity}
\end{equation}
Since $|r_j(f)-\kappa_j|\le\lVert f\rVert/\tau$, the learning score obeys
\[
 |q_j(f)-\widehat q_j(f)|\le\frac{\lVert f\rVert}{\tau\nu^3}.
\]
At fixed class states, this guarantee also propagates through the two-view learning objective without a multiplicative dependence on the number of classes or the long-tail priors. Let $\delta_i^{(v)}=\lVert f_i^{(v)}\rVert/(\tau\nu^3)$ and $\overline\delta^{(v)}=B^{-1}\sum_i\delta_i^{(v)}$. Replacing $q_j$ by $\widehat q_j$ then gives
\begin{equation}
 |\widehat L_{\mathrm{PATT}}-L_{\mathrm{PATT}}|
 \le\overline\delta^{(2)}+\overline\delta^{(3)},
 \label{eq:main-learning-certificate}
\end{equation}
for arbitrary additive class biases. Corresponding probability, margin, feature-gradient, and class-group bounds accompany the certificate in Appendix~\ref{sec:patt-theory} and are proved in Appendix~\ref{app:patt-proof}.

Finite recurrence follows an order-dependent chain, whereas AR/FR contains a
fixed number of elementary operations. With $p=2\nu+2$, the certificate becomes
$\nu^{-3}=8/(p-2)^3$. This yields a useful high-dimensional reversal:
increasing feature dimension makes the finite recurrence sequentially deeper,
while the worst-case AR/FR approximation guarantee becomes tighter. Thus the
regime in which recurrence becomes increasingly expensive is also the regime
in which the paired analytic approximation is most strongly certified. AR/FR
therefore combines coherence, certified fidelity, and fixed-depth computation
in the high-dimensional setting that motivates the method.

\section{Experiments}
\label{sec:experiments}
\subsection{Experimental setup and numerical realizations}
We compare three numerical realizations: \emph{original recurrence} uses the finite forward and clipped backward; \emph{consistent recurrence} differentiates that same forward; and \emph{AR/FR} replaces the potential and derivative jointly. The first comparison isolates the backward rule, while AR/FR tests a paired replacement. Architecture, class-state estimation, data, optimizer, and surrounding objective are matched; implementation and training details appear in Appendices~\ref{app:current-protocol} and~\ref{sec:system}.

\subsection{Same forward, different gradients and trajectories}
\begin{figure}[htbp]\centering
\includegraphics[width=0.80\linewidth]{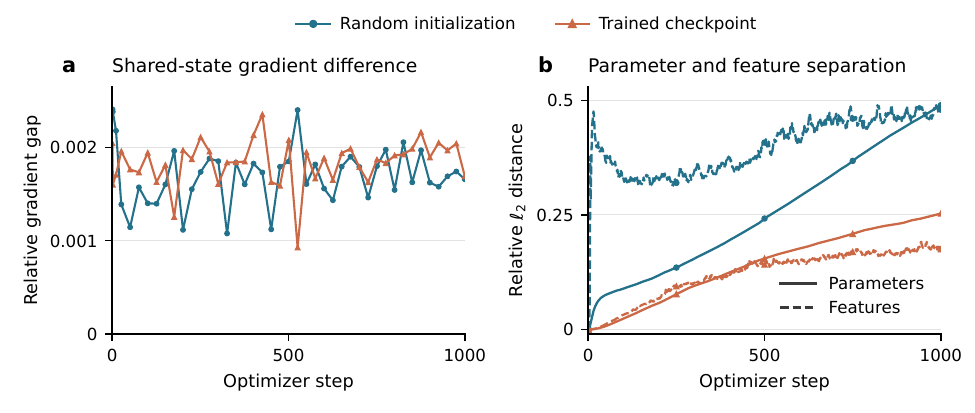}
\caption{\textbf{Same-state gradient mismatch becomes trajectory separation.}
Original and consistent recurrence share random or intermediate-training starting states and augmented inputs. Left: local gradient gaps with bitwise-identical forward values. Right: parameter and feature distances after separate updates.}
\label{fig:patt-trajectory}
\end{figure}

Changing only the backward rule alters PATT optimization from both random initialization and an intermediate checkpoint. Within each fork, original and consistent recurrence share initial parameters, class and optimizer states, and augmented inputs. Shared-state comparisons yield bitwise-identical scores and losses but different feature gradients. Over 1000 updates, these local differences accumulate into persistent parameter and representation separation (Figure~\ref{fig:patt-trajectory}), linking numerical incoherence to the realized learning dynamics.

\subsection{When does mismatch occur?}
\label{sec:mismatch-regimes}
\begin{figure}[H]\centering
\includegraphics[width=0.94\linewidth]{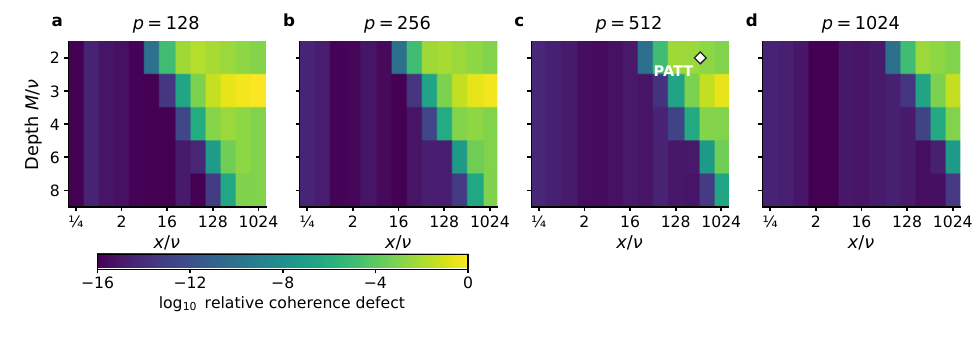}
\caption{\textbf{Concentration and depth jointly determine coherence.}
Colour shows $\log_{10}(|g_{\rm supplied}-\widetilde\Phi'|/|\widetilde\Phi'|)$, with a $10^{-16}$ display floor. The diamond locates captured PATT queries at $p=512$ and $M/\nu=2$.}
\label{fig:proco-conditions}
\end{figure}

Coherence defects depend jointly on concentration and recurrence depth (Figure~\ref{fig:proco-conditions}). At high concentration, deeper recurrence does not uniformly reduce the clipped defect: improving the raw ratio and matching the finite-forward derivative are distinct objectives. Captured PATT queries occupy this high-concentration regime. ProCo exhibits the same conditional phenomenon in a second vMF implementation (Appendix~\ref{app:proco-extended}).

\FloatBarrier
\Needspace{22\baselineskip}
\subsection{Fidelity of AR/FR}
\begin{wrapfigure}{r}{0.40\textwidth}
\centering
\includegraphics[width=\linewidth]{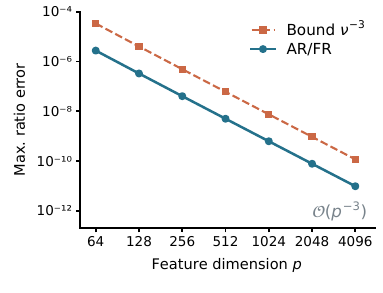}
\caption{\textbf{Cubic error decay.} Blue: sampled maximum error. Orange: the all-axis certificate $\nu^{-3}$.}
\label{fig:dimension-error}
\end{wrapfigure}
Across feature dimensions from 64 to 4096, the maximum sampled AR/FR ratio error decays with fitted log--log slope $-3.02$ and remains below the uniform certificate (Figure~\ref{fig:dimension-error}). This cubic trend supports the high-dimensional behavior established in Section~\ref{sec:high-dimension}: fidelity improves as the recurrence being replaced grows deeper. Captured-state score errors, potential-difference errors, and evaluator comparisons appear in Appendix~\ref{app:non-timing}.

\subsection{Complete task performance}
We evaluate classification and SC-OOD detection on CIFAR-10-LT and CIFAR-100-LT at imbalance factors 10, 50, and 100, together with ImageNet-LT. Table~\ref{tab:task-summary} reports accuracy and AUROC; Appendix~\ref{app:aggregate-results} includes FPR95, and Appendix~\ref{app:current-protocol} gives the evaluation protocol.

The paired replacement preserves broadly comparable task performance, with changes that depend on the dataset, imbalance, and metric. On CIFAR, accuracy and OOD detection can move in different directions, and the most imbalanced CIFAR-100 setting shows modest decreases. ImageNet-LT shows improvements across all three metrics. These results characterize the task behavior of numerical replacement across settings.

\begin{table}[!htbp]\centering\small
\caption{\textbf{Classification and OOD detection under different numerical realizations.} CIFAR results report mean $\pm$ standard deviation over three runs. Metrics are percentages; IF denotes the imbalance factor. Full FPR95 results appear in Appendix~\ref{app:aggregate-results}. Bold and underline indicate the best and second-best values within each setting.}
\label{tab:task-summary}
\setlength{\tabcolsep}{4pt}
\begin{tabular}{@{}ll l ccc@{}}\toprule
Dataset & IF & Metric & Original recurrence & Consistent recurrence & AR/FR\\\midrule
CIFAR-10-LT & 10 & Acc. $\uparrow$ & $92.04\pm 0.17$ & $\underline{92.12}\pm 0.18$ & $\mathbf{92.16}\pm 0.19$ \\
 &  & AUROC $\uparrow$ & $\underline{93.07}\pm 0.13$ & $92.30\pm 0.27$ & $\mathbf{93.43}\pm 0.46$ \\
\midrule
CIFAR-10-LT & 50 & Acc. $\uparrow$ & $87.30\pm 0.29$ & $\mathbf{87.58}\pm 0.35$ & $\underline{87.42}\pm 0.15$ \\
 &  & AUROC $\uparrow$ & $90.18\pm 2.46$ & $\underline{90.95}\pm 1.57$ & $\mathbf{91.02}\pm 1.13$ \\
\midrule
CIFAR-10-LT & 100 & Acc. $\uparrow$ & $83.82\pm 0.21$ & $\underline{84.67}\pm 0.41$ & $\mathbf{84.73}\pm 0.31$ \\
 &  & AUROC $\uparrow$ & $\mathbf{90.66}\pm 0.13$ & $\underline{89.97}\pm 0.39$ & $89.95\pm 0.34$ \\
\midrule
CIFAR-100-LT & 10 & Acc. $\uparrow$ & $\underline{62.37}\pm 2.43$ & $62.12\pm 1.84$ & $\mathbf{62.56}\pm 2.03$ \\
 &  & AUROC $\uparrow$ & $\underline{78.18}\pm 2.09$ & $78.08\pm 1.23$ & $\mathbf{78.44}\pm 1.53$ \\
\midrule
CIFAR-100-LT & 50 & Acc. $\uparrow$ & $\underline{53.26}\pm 0.19$ & $52.87\pm 0.31$ & $\mathbf{53.74}\pm 0.31$ \\
 &  & AUROC $\uparrow$ & $75.37\pm 0.68$ & $\underline{75.54}\pm 0.60$ & $\mathbf{76.14}\pm 0.24$ \\
\midrule
CIFAR-100-LT & 100 & Acc. $\uparrow$ & $\mathbf{50.33}\pm 0.16$ & $49.48\pm 0.26$ & $\underline{49.66}\pm 0.16$ \\
 &  & AUROC $\uparrow$ & $\mathbf{75.78}\pm 0.33$ & $\underline{75.40}\pm 0.54$ & $75.21\pm 0.65$ \\
\midrule
ImageNet-LT & --- & Acc. $\uparrow$ & $53.88$ & $\underline{54.11}$ & $\mathbf{54.34}$ \\
 &  & AUROC $\uparrow$ & $76.19$ & $\underline{76.87}$ & $\mathbf{77.50}$ \\
\bottomrule
\end{tabular}
\end{table}

\subsection{Efficient realization and structure-aware execution}
We evaluate the complete local vMF path under a unified execution protocol, covering class-state updates, score geometry, cross-entropy, and feature backward while excluding the backbone and data pipeline.

\Needspace{23\baselineskip}
\noindent\begin{minipage}[t]{0.49\textwidth}
\vspace{0pt}
Figure~\ref{fig:efficiency-compact} compares the numerical realizations and the structure-aware implementation. Consistent recurrence restores coherence by differentiating the finite forward but increases both runtime and memory. Log-domain Miller retains recurrence in a different numerical representation. AR/FR alone reduces runtime by $54.4\times$ with a similar dense-layout memory footprint. Adding direct class-statistic accumulation, factorized geometry, and compiled GPU execution yields a further $23.8\times$ runtime reduction. The overall $1293\times$ speedup and $15.8\times$ memory reduction characterize this combined realization; the memory savings arise from structure-aware execution. Runtime and incremental peak memory are measured over the same local execution boundary for all realizations; the detailed workload, memory definition, and measurement protocol are provided in Appendix~\ref{app:h100-local}.
\end{minipage}\hfill
\begin{minipage}[t]{0.47\textwidth}
\vspace{0pt}
\centering
\includegraphics[width=\linewidth]{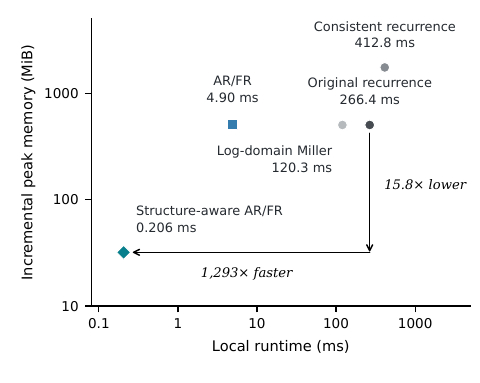}
\captionof{figure}{\textbf{Local runtime--memory trade-off.}}
\label{fig:efficiency-compact}
\end{minipage}

\section{Discussion}
The controlled recurrence intervention shows that the supplied derivative is part of the learning algorithm: an unchanged forward objective can lead to different optimization trajectories. AR/FR pairs a coherent derivative with a certified forward approximation, preserving broadly comparable task performance while enabling fixed-depth execution. The Jacobian construction establishes a local mechanism rather than its prevalence across training states.

More broadly, the vMF case separates two questions that are often collapsed in
differentiable numerical software. Approximation analysis asks whether a
numerical value or derivative is close to an analytic reference; coherence asks
whether the value and derivative supplied by the program belong to each other.
The latter matters even when both individual errors are small, because
optimization acts on their composition. Pairing the realized scalar with its
derivative therefore provides a useful design and diagnostic principle for
learning objectives built from nontrivial numerical primitives.

\section{Conclusion}
Finite numerical evaluation can preserve the forward objective while changing gradients and optimization trajectories. We characterize this decoupling in vMF learning and its effect on local update-field geometry. AR/FR pairs the potential and derivative to combine coherence, certified fidelity, and fixed-depth execution. Differentiable numerical primitives should be evaluated through both their realized values and supplied derivatives.

\paragraph{AI Use Statement.}
Generative AI assisted with writing and editing, feedback on theoretical derivations, experiment design and result interpretation, drafts of mathematical checks, code development, experiment organization, and figure preparation. The authors independently verified all AI-assisted mathematical claims, code, analyses, and written content against derivations, executable code, primary references, and raw experimental records, and take full responsibility for the final content.

\paragraph{Reproducibility Statement.}
Appendix~\ref{app:current-protocol} describes the training and evaluation protocols, and Appendix~\ref{app:intervention-protocol} specifies the controlled trajectory interventions. Theoretical proofs and numerical validation are provided in Appendices~\ref{app:theory} and~\ref{app:numerical-validation}, with the local runtime and memory protocol in Appendix~\ref{app:h100-local}.

\bibliographystyle{arxiv_references}
\bibliography{references}

\clearpage
\appendix
\begingroup
\begin{center}
{\Large\bfseries SAME LOSS, DIFFERENT GRADIENTS\par}
\medskip
{\Large Supplementary Material\par}
\end{center}

\medskip
{\large\bfseries Contents\par}

\definecolor{contentsblue}{RGB}{47,112,171}
\hypersetup{linkcolor=contentsblue}
\setcounter{tocdepth}{2}

\makeatletter
\renewcommand*\l@section[2]{%
  \vspace{4pt}%
  \@dottedtocline{1}{0em}{1.5em}{\bfseries #1}{\bfseries #2}}
\renewcommand*\l@subsection[2]{%
  \@dottedtocline{2}{1.5em}{2.3em}{#1}{#2}}

\small
\newif\ifshowappendixcontents
\showappendixcontentsfalse

\let\originalcontentsline\contentsline
\def\appendixstart{appendix.A}
\def\appendixstartalt{section.A}

\renewcommand\contentsline[4]{%
  \def\currentanchor{#4}%
  \ifx\currentanchor\appendixstart
    \showappendixcontentstrue
  \fi
  \ifx\currentanchor\appendixstartalt
    \showappendixcontentstrue
  \fi
  \ifshowappendixcontents
    \originalcontentsline{#1}{#2}{#3}{#4}%
  \fi}

\@starttoc{toc}
\makeatother
\endgroup

\clearpage
\FloatBarrier
\section{Related work}
\label{app:related}

\paragraph{Supplied gradients and numerical differentiation.}
The possibility of using a backward rule different from the forward
derivative is established, including intentional straight-through estimators
for stochastic or nonsmooth units \citep{bengio2013estimating,yin2019understanding}.
Automatic differentiation evaluates derivatives of numerical programs
\citep{baydin2018automatic}; nonsmooth models and conservative set-valued
fields clarify the mathematical interpretation of the resulting oracles
\citep{bolte2020model,bolte2021conservative}. These nonsmooth frameworks
provide context for distinguishing a supplied update from a classical gradient.
Our setting concerns smooth special-function objectives whose numerical
forward and backward are intended to represent an analytic value--derivative
relation. The contribution is a characterization of a concrete recurrence
mechanism, its composition and training consequences, and a certified paired
replacement.

\paragraph{vMF probabilistic learning.}
Directional clustering uses vMF mixtures to model normalized observations
\citep{banerjee2005clustering}, while hyperspherical VAEs use vMF latent
distributions \citep{davidson2018hyperspherical}. Differentiating through
probability distributions also motivates implicit reparameterization
\citep{figurnov2018implicit}; that work concerns sampling gradients rather
than the finite normalizer recurrence examined here.
ProCo derives scores from class-conditional vMF distributions
\citep{du2024proco}, and PATT uses the score in long-tailed OOD learning
\citep{he2025patt}. They provide related implementations with different
visited numerical states. The supervised vMF loss of
\citet{Scott_2021_ICCV} provides a complementary construction: it averages
analytic ratio bounds and integrates that approximation to obtain a
normalizer surrogate. The ratio and potential are therefore paired already
at the mathematical level. Such a coherent reference separates the value
of pairing from the particular approximation and certificate developed here.
SC-OOD evaluates novelty after accounting for semantic overlap
\citep{yang2021scood}.

\paragraph{Contrastive representations.}
SimCLR learns representations from augmented positive pairs
\citep{chen2020simple}, and supervised contrastive learning incorporates
class labels into the positive-pair construction \citep{khosla2020supervised}.
Alignment and uniformity characterize complementary properties of contrastive
representations on the hypersphere \citep{wang2020alignment}. ProCo moves
from sampled pairs to class distributions; our study addresses the numerical
realization of the resulting probabilistic score rather than changing the
pair construction or representation objective.

\paragraph{Bessel computation and approximation.}
Classical recurrence and scaling methods \citep{amos1974computation,sra2012note},
analytic ratio bounds \citep{hornik2013amos,ruizantolin2016bounds,segura2023simple},
and uniform large-order expansions \citep{olver1954asymptotic} underpin
high-order special-function evaluation. CUSF supplies accurate GPU
log-Bessel computation \citep{Plesner_2024} and is a relevant numerical and
systems comparator. Our AR/FR pair builds on the classical asymptotic
structure; its contribution comprises the paired realization, uniform
certificate, and connection to the recurrence mechanisms studied here.
The numerical evaluation reports regimes favoring different evaluators.

\paragraph{Long-tailed OOD context.}
Long-tailed recognition can be addressed through class-dependent margins
\citep{cao2019ldam}, decoupled representation and classifier learning
\citep{kang2020decoupling}, or label-frequency-based logit adjustment
\citep{menon2021logit}. In OOD detection, maximum Softmax probability
\citep{DBLP:conf/iclr/HendrycksG17} and energy scores
\citep{NEURIPS2020_f5496252} provide scoring rules, while outlier exposure
uses auxiliary outliers during training \citep{DBLP:conf/iclr/HendrycksMD19}.
PATT combines vMF-based implicit semantic augmentation with tail-focused
calibration \citep{he2025patt}. Existing LT-OOD approaches alter training
objectives, representation separation, or outlier modeling
\citep{wei2024eat,pmlr-v162-wang22aq,Miao_2024}; further approaches use
outlier-distribution adaptation \citep{NEURIPS2024_ee657707} or inter-sample
graph information \citep{Udayangani_2025}. Our comparison holds the host
learning and calibration pipeline fixed to study its numerical realization.

\section{Experimental protocols and reproducibility}
\subsection{Training and evaluation protocols}
\label{app:current-protocol}

\paragraph{CIFAR training and readout.}
The CIFAR datasets originate from \citet{krizhevsky2009learning}.
The complete three-realization comparison uses exponential long-tail
subsampling with factors 10, 50, and 100. CIFAR results report means and standard deviations over three runs, with matched initialization across realizations within each run. The shared training
shell uses ResNet18 \citep{he2016deep}, ID batch 128 with three views, OE batch 256 from the
300k Tiny Images source, Adam, and the original augmentation pipeline.
FP32 backbone and expanded geometry and FP64 scalar evaluation are shared
across branches; autocast is off. The final epoch-99 weights are evaluated,
without selecting a best checkpoint. All branches use matched training configurations and common initialization; the causal forks additionally share the exact augmented tensors at every update.

Evaluation applies PATT feature calibration fitted using only members of
the corresponding long-tailed training split. The test set is reserved for
evaluation. SC-OOD averages source-level metrics after semantic-ID handling
within each of six sources.

\paragraph{ImageNet-LT training and evaluation.}
The ImageNet-LT realizations use ResNet50, 100 epochs, SGD with initial
learning rate 0.1, momentum 0.9 and weight decay $5\times10^{-5}$, ID batch
32 and OE batch 64. Final checkpoints have zero-based epoch 99.
Calibration uses deterministic class-balanced sampling from the training
ID data and training-OOD data, with 30,000 calibration samples per source;
held-out labels are used only for evaluation. Evaluations share
manifests, transforms, seed and batch size. Classification uses the
20,000-image ImageNet-LT validation manifest and OOD detection uses the
50,000-image test-OOD manifest.

\subsection{Controlled intervention protocols}
\label{app:intervention-protocol}
\paragraph{PATT causal forks.}
Random-initialization and trained-checkpoint forks each run 1000 updates
from a shared state within each fork. The second fork starts from the
original branch after its first 1000 updates. Both branches have identical initial parameters, class banks, optimizer settings, and
actual augmented tensors. Local probes evaluate different backward rules at
one shared state; evolved-state measurements evaluate each branch after
its own updates. Each fork contains 43 local probes with bitwise-equal
original and consistent-recurrence forward values. Across the 43 probes, relative gradient differences remain on the order of $10^{-3}$ for both starting states. Relative evolved parameter distance
is normalized by the original parameter norm. Both forks develop substantial parameter separation over 1000 updates.

\subsection{Source implementation}
The original-recurrence baseline follows the public PATT implementation used throughout our experiments, with Miller start $M=2\nu$ and a unit-clipped custom backward. Exact revision information is provided with the submission artifacts.

\section{ProCo: state-dependent coherence defects}
\label{app:proco-extended}
We use ProCo as a second real implementation of finite-recurrence vMF learning. The audit compares its supplied clipped radial rule with FP64 automatic differentiation of the same finite forward at $M=2\nu$, testing coherence at fixed numerical states.

\paragraph{Observed states.}
The CIFAR100-LT pure-ProCo captures use ResNet32, $p=128$, $\tau=0.1$, and imbalance factor 100. Each first-batch capture contains 25,600 query radii; the end-of-epoch-1 replay contains 51,200. The ImageNet joint-training replay uses an epoch-90 checkpoint with $p=1024$ and a new augmented batch, yielding 64,000 radii. The iNaturalist audit covers 8,142 stored class-bank concentrations at each of epochs 1, 10, 50, and 90. These bank values are reported separately from query radii.

\begin{figure}[!htbp]\centering
\includegraphics[width=\linewidth]{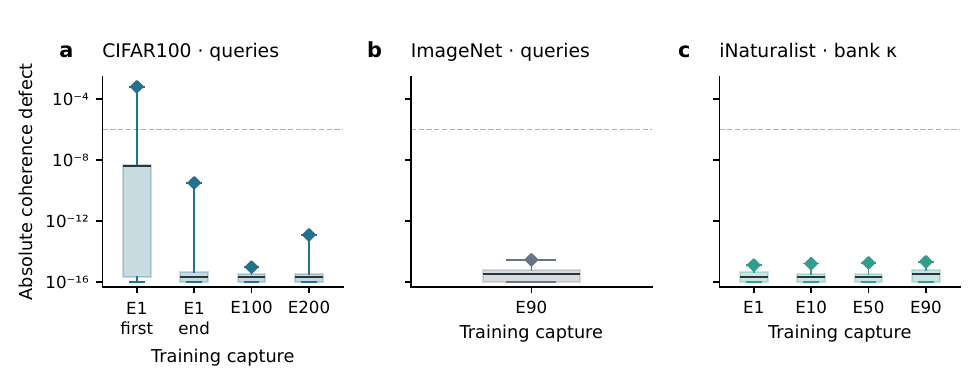}
\caption{\textbf{ProCo coherence defects depend on the visited state.}
Absolute differences compare the supplied radial rule with FP64 autodiff at $M=2\nu$. Boxes show median and interquartile range; whiskers span extrema. (a) CIFAR queries, (b) ImageNet queries, and (c) iNaturalist class-bank concentrations. The dashed diagnostic scale is $10^{-6}$; values below $10^{-16}$ share a display floor.}
\label{fig:proco-state-distribution}
\end{figure}

\paragraph{State dependence.}
In the first CIFAR batch, $23\%$ of query radii have absolute coherence defect above $10^{-6}$. By epochs 100 and 200, the corresponding captures are near floating-point agreement; the ImageNet replay and iNaturalist bank audits likewise have small defects (Figure~\ref{fig:proco-state-distribution}). Early CIFAR states include capped concentration estimates from singleton classes. The observed defect thus depends on the regime reached by the class estimator and query geometry, rather than being uniformly large throughout training.

\section{Theoretical proofs and auxiliary identities}
\label{app:theory}
\subsection{Learning-level fidelity guarantees}
\label{sec:patt-theory}

The central question is whether classes or long-tail priors amplify FR error.
Class priors enter as fixed additive biases $b_j=\log\pi_j$. For view $v\in\{2,3\}$, set
$\delta_i^{(v)}=\norm{f_i^{(v)}}/(\tau\nu^3)$ and
$\overline\delta^{(v)}=B^{-1}\sum_i\delta_i^{(v)}$.

\begin{theorem}[Fixed-state PATT evaluator certificate]
\label{thm:patt}
Assume $\nu\ge10/9$ and $\tau>0$. Fix each view's online class state, let $K<\infty$ and additive biases $b_j$ be
arbitrary, and replace only $q_j$ by \eqref{eq:f2-score}.  Then
\begin{align}
 |q_j(f_i^{(v)})-\widehat q_j(f_i^{(v)})|
 &\le\delta_i^{(v)},
 \label{eq:patt-score-bound}\\
 |\widehat L_{\mathrm{PATT}}-L_{\mathrm{PATT}}|
 &\le\overline\delta^{(2)}+\overline\delta^{(3)}.
 \label{eq:patt-total-bound}
\end{align}
The bounds have no multiplicative factor in the class count, priors, or
imbalance ratio.
Appendix~\ref{app:patt-proof} gives the proof and the corresponding Softmax-TV,
margin, feature-gradient, and nonempty-group bounds.
\end{theorem}

The class-count independence follows from log-sum-exp monotonicity under a
uniform coordinate perturbation, using the maximum coordinate error.  It accommodates PATT's
$b_j=\log\pi_j$ even under severe imbalance.

The same argument controls adjusted probabilities and feature gradients at
fixed class states. The learning-level guarantee therefore connects the
special-function approximation to the quantities consumed by optimization.
\subsection{Proof of the Softmax integrability obstruction}
\label{app:curl-proof}
Since $w_j=\nabla_f H(r_j)$, its Jacobian is symmetric.
Also $\nabla_f p_j=p_j(v_j-\bar v)$. Differentiating $G$ and subtracting
its transpose leaves
\[
 \sum_jp_j(w_jv_j^\top-v_jw_j^\top)
 -\bar w\bar v^\top+\bar v\bar w^\top.
\]
For every class, $w_j$ and $v_j$ are collinear with $\nabla_f r_j$,
so the sum is zero. Symmetry of the Hessian is necessary for a $C^2$
potential, proving the obstruction. When $g=F'$, the ordinary chain rule
gives the final assertion. Hence the condition provides a sufficient obstruction; incoherence alone does not imply nonconservativity.
\hfill$\square$

\paragraph{A reproducible two-class construction.}
Take $p=128$, $\nu=63$, $M=126$, $\tau=1$, $f=0$, zero class biases,
$\mu_1=e_1$, $\mu_2=e_2$, and $(\kappa_1,\kappa_2)=(4032,8064)$.
Here $F=\widetilde\Phi_{63,126}$ and $g=\min\{\widetilde R_{63,126},1\}$.
Both scores are zero, so both probabilities are $1/2$.
The raw ratios are approximately 1.08748 and 1.56080, placing both classes
strictly inside a smooth clipping region. The finite-forward derivatives
are approximately 0.984317 and 0.992073. With the convention
$J_{ij}=\partial G_i/\partial f_j$, direct substitution yields
\[
 J_{12}-J_{21}
 =\frac{F'(\kappa_1)g(\kappa_2)-g(\kappa_1)F'(\kappa_2)}{4\tau^2}
 =-1.93915\times10^{-3}.
\]
The true derivative of the same forward produces zero antisymmetry.
Class states are fixed; the local smoothness assertion excludes clipping
transition points and zero query radii.

\paragraph{Degeneracy and controlled mismatch.}
If $g=cF'$ with the same constant $c$ throughout a neighborhood, then
$G=c\nabla_f L$ is conservative even when $c\ne1$. Nonzero coherence defect
alone therefore does not imply nonconservativity. At the origin of the
orthogonal two-class construction, equal concentrations make the displayed
antisymmetry zero; this pointwise statement does not assert integrability
throughout a neighborhood.
For a constant intervention strength $\lambda$, define
$g_\lambda=F'+\lambda(g-F')$. Since the forward probabilities are unchanged,
\[
 G_\lambda-G_0=\lambda(G_1-G_0),\qquad
 J_{G_\lambda}-J_{G_\lambda}^{\top}
 =\lambda(J_{G_1}-J_{G_1}^{\top}).
\]
The linear relation holds for the fixed-state field on smooth regions.

\subsection{PATT objective and finite-start identities}
\label{app:released-details}

\subsubsection{The vMF term inside PATT}

Let $p=2\nu+2$ be the feature dimension.  For class $j$, let $\mu_j$ be a
unit direction, $\kappa_j\ge0$ a concentration, $f\in\mathbb R^p$ a query
feature, and $\tau>0$ the temperature.  Define
\begin{equation}
 r_j(f)=\norm{\kappa_j\mu_j+f/\tau},\qquad
 \phinu(x)=\log I_\nu(x)-\nu\log x.
 \label{eq:patt-score-def}
\end{equation}
At $x=0$, $\phinu$ denotes its continuous extension; finite-start sign and
parity statements concern $x>0$.  The exact class score is
\begin{equation}
 q_j(f)=\phinu(r_j(f))-\phinu(\kappa_j).
 \label{eq:exact-score}
\end{equation}
PATT adds a fixed long-tail log prior $b_j=\log\pi_j$ before cross-entropy.
For views $f_i^{(2)},f_i^{(3)}$ with label $y_i$, write
\begin{equation}
 \ell_i^{(v)}=-q_{y_i}(f_i^{(v)})-b_{y_i}
 +\log\sum_{j=1}^{K}\exp(q_j(f_i^{(v)})+b_j).
 \label{eq:view-loss}
\end{equation}
For a batch of size $B$, the implicit semantic augmentation term is
\begin{equation}
 L_{\mathrm{ISAC}}=\frac{1}{2B}\sum_{i=1}^{B}
 \left(\ell_i^{(2)}+\ell_i^{(3)}\right).
 \label{eq:isac}
\end{equation}
The complete original objective has the form
\begin{equation}
 L_{\mathrm{PATT}}=L_{\mathrm{ISAC}}
 +\lambda_{\mathrm{TLA}}L_{\mathrm{TLA}}
 +\lambda_{\mathrm{OE}}L_{\mathrm{OE}}.
 \label{eq:patt-total}
\end{equation}
Only $L_{\mathrm{ISAC}}$ contains the Bessel score.  The prior bias, TLA term,
OE term, encoder, data, and feature-calibration evaluator do not depend on the
choice between Miller and FR.

\subsubsection{Finite-sample class state before Bessel evaluation}

For class count $n_j>0$ and detached-feature sum $S_j$, PATT's cumulative-mean
estimator uses
\begin{equation}
 a_j=\frac{S_j}{n_j},\qquad R_j=\norm{a_j},\qquad
 \mu_j=\frac{a_j}{R_j},\qquad
 \kappa_j=h_p(R_j)=\frac{pR_j}{1-R_j^2}.
 \label{eq:released-state}
\end{equation}
When $R_j=0$, we use the continuous score-geometry convention
$\kappa_j=0$ and $\kappa_j\mu_j=0$ (equivalently, set $\mu_j=0$); no result
depends on an otherwise undefined direction.  The source initializer handles
$n_j=0$ separately.
Appendix Lemma~\ref{lem:finite-state} characterizes the finite-sample resultant.  The source keeps an initializer at $n_j=0$
and caps the raw concentration at $10^5$.  For batch sum $s$ and count $m$,
Appendix Proposition~\ref{prop:state-reduction} gives the exact transition
$a^+=(Na+s)/(N+m)$, preserving both estimator objects and the sequential,
possibly correlated view schedule without a dense one-hot expansion.

\subsubsection{The original forward--backward objects}

The exact exponential-family geometry links value and derivative:
\begin{equation}
 \phinu'(x)=\Rnu(x),\qquad
 \Rnu(x)=\frac{I_{\nu+1}(x)}{I_\nu(x)}.
 \label{eq:potential-identity}
\end{equation}
The original evaluator starts Miller's order recurrence at $M=2\nu$.  For a
general integer $M>\nu\ge1$, define
\begin{equation}
 b_M=1,\quad b_{M+1}=0,\quad
 b_{i-1}=\frac{2i}{x}b_i+b_{i+1}\quad(i=M,\ldots,1).
 \label{eq:miller-b}
\end{equation}
It is useful first to isolate the exact finite-recurrence object
\begin{align}
 \widetilde\Phi_{\nu,M}(x)
 &=\log I_0(x)+\log\frac{b_\nu}{b_0}-\nu\log x,
 \label{eq:miller-potential}\\
 \widetilde R_{\nu,M}(x)&=\frac{b_{\nu+1}}{b_\nu},
 \qquad
 \overline R_{\nu,M}(x)=\min\{\widetilde R_{\nu,M}(x),1\}.
 \label{eq:miller-three}
\end{align}
The source uses $\log(x+\epsilon)$ with $\epsilon=10^{-20}$, saves
$\widetilde R_{\nu,M}$, and supplies $\overline R_{\nu,M}$ in backward;
Appendix~\ref{app:miller-score-proof} retains the literal correction.

\paragraph{Potential-consistent evaluator.}
A reported forward $\Psi$ and supplied radial backward $G$ are
\emph{potential-consistent} on $I$ when $\Psi'=G$ throughout $I$. The corresponding
signed residual is $D_g^\epsilon=(\widetilde\Phi_{\nu,M}^{\epsilon})'-g=-d^\epsilon$, the negative of the coherence defect in Section~\ref{sec:realization}.

\subsubsection{Finite-start parity is an all-axis effect}

Set
\begin{equation}
 \widetilde u_i=\frac{b_{i+1}}{xb_i},\qquad
 u_i=\frac{I_{i+1}(x)}{xI_i(x)},\qquad
 T_i(u)=\frac{1}{2i+x^2u}.
 \label{eq:miller-u}
\end{equation}
Then $\widetilde u_M=0$ and both sequences obey
$\widetilde u_{i-1}=T_i(\widetilde u_i)$ and
$u_{i-1}=T_i(u_i)$.

\begin{theorem}[Finite-start sign and product]
\label{thm:miller-sign}
For every integer $M>\nu\ge1$ and $x>0$,
\begin{equation}
\begin{split}
 \widetilde u_\nu-u_\nu
 ={}&(-1)^{M-\nu}(\widetilde u_M-u_M)\\
 &\times\prod_{i=\nu+1}^{M}
 \frac{x^2}
 {(2i+x^2\widetilde u_i)(2i+x^2u_i)}.
\end{split}
\label{eq:miller-product}
\end{equation}
Consequently,
\begin{equation}
 \operatorname{sign}(\widetilde R_{\nu,M}-\Rnu)
 =(-1)^{M-\nu+1}.
 \label{eq:miller-sign}
\end{equation}
At the original start $M=2\nu$, the sign is $(-1)^{\nu+1}$.
\end{theorem}

All factors in \eqref{eq:miller-product} are positive, so the theorem is an
exact-arithmetic statement over the entire positive axis.  Intuitively, every backward map $T_i$ is strictly
decreasing on the positive axis: the terminal approximation starts below the
exact tail, and each recurrence step flips the inequality.  Only the number of
steps from $M$ to $\nu$ determines which side the final ratio occupies.  PATT
uses $\nu=63$ at $p=128$ and $\nu=255$ at
$p=512$; both are odd.  Hence their raw finite-start ratios strictly exceed
the exact Bessel ratio for every $x>0$.

\begin{theorem}[High-concentration parity phase]
\label{thm:miller-parity}
For $M=2\nu$ and $x\to\infty$,
\begin{align}
 \widetilde R_{\nu,2\nu}(x)
 &=\frac{2x}{(\nu+1)(3\nu+1)}+O(x^{-1})\to\infty
 &&(\nu\text{ odd}),
 \label{eq:miller-odd-asymptotic}\\
 \widetilde R_{\nu,2\nu}(x)
 &=\frac{\nu(3\nu+2)}{2x}+O(x^{-3})\to0
 &&(\nu\text{ even}),
 \label{eq:miller-even-asymptotic}\\
 \widetilde\Phi_{\nu,2\nu}(x)
 &=x-(\nu+3/2)\log x+C_\nu+O(x^{-1}),
 \label{eq:miller-forward-asymptotic}\\
 \widetilde\Phi_{\nu,2\nu}'(x)
 &=1-\frac{\nu+3/2}{x}+O(x^{-2}).
 \label{eq:miller-forward-derivative-asymptotic}
\end{align}
where the last two identities hold for odd $\nu$.
\end{theorem}

The raw ratio therefore diverges for the odd orders used by PATT, while both
the finite-forward derivative and $\Rnu=1-(\nu+1/2)/x+O(x^{-2})$ approach
one.  By continuity, the raw ratio crosses one at least once.  The concentration scan in Section~\ref{sec:experiments} evaluates this crossing behavior.

Since $p=2\nu+2$, every $p\equiv0\pmod4$ gives odd $\nu$ and hence all-axis
overshoot plus high-concentration divergence; this includes $p=128,512$.

\subsubsection{Clipping is not a potential repair}

Define the finite-potential error and the two idealized $\epsilon=0$
signed residuals by
\begin{equation}
\begin{split}
 E_{\nu,M}&=\widetilde\Phi_{\nu,M}-\phinu,\\
 D_{\mathrm{raw}}&=\widetilde\Phi_{\nu,M}'-\widetilde R_{\nu,M},\\
 D_{\mathrm{clip}}&=\widetilde\Phi_{\nu,M}'-\overline R_{\nu,M}.
\end{split}
\label{eq:defects}
\end{equation}
Since $\phinu'=\Rnu$,
\begin{equation}
 D_{\mathrm{raw}}=E_{\nu,M}'-(\widetilde R_{\nu,M}-\Rnu),
 \quad
 D_{\mathrm{clip}}=E_{\nu,M}'-(\overline R_{\nu,M}-\Rnu).
 \label{eq:defect-decomposition}
\end{equation}
For the ideal finite-forward score
$\widetilde q_j=\widetilde\Phi_{\nu,M}(r_j)
-\widetilde\Phi_{\nu,M}(\kappa_j)$, the exact identity is
\begin{equation}
 \boxed{\widetilde q_j-q_j
 =E_{\nu,M}(r_j)-E_{\nu,M}(\kappa_j).}
 \label{eq:miller-score-endpoint}
\end{equation}
Appendix~\ref{app:miller-score-proof} gives the literal $\epsilon$ correction,
raw/clipped integral identities, and the logarithmic high-concentration
finite-forward score error.  Table~\ref{tab:captured-audit} contrasts captured-state score fidelity with backward coherence.

\subsection{Proof of the finite-start sign theorem}
\label{app:miller-sign-proof}

For $T_i(u)=(2i+x^2u)^{-1}$, direct subtraction gives the one-step
M\"obius identity
\begin{equation}
 T_i(a)-T_i(b)
 =-\frac{x^2(a-b)}
 {(2i+x^2a)(2i+x^2b)}.
 \label{eq:mobius-step-app}
\end{equation}
The denominators are strictly positive for the recurrence sequences in
\eqref{eq:miller-u}.  Apply \eqref{eq:mobius-step-app} successively for
$i=M,M-1,\ldots,\nu+1$.  Each step contributes one negative sign and one
positive factor, yielding exactly \eqref{eq:miller-product}.  Because the
original terminal is $\widetilde u_M=0$ while
$u_M=I_{M+1}(x)/(xI_M(x))>0$, multiplication by $x>0$ gives
\eqref{eq:miller-sign}.  Substituting $M=2\nu$ completes the proof of
Theorem~\ref{thm:miller-sign}.

\subsection{High-concentration asymptotics}
\label{app:miller-parity-proof}

Write $y=x^2$.  Starting from the even index $M=2\nu$, the finite recurrence
alternates between
\begin{equation}
 \widetilde u_j=\frac{c_j}{y}+O(y^{-2})
 \quad\text{at even }j,
 \qquad
 \widetilde u_j=a_j+O(y^{-1})
 \quad\text{at odd }j,
 \label{eq:parity-ansatz-app}
\end{equation}
where $a_j,c_j>0$ except for the terminal coefficient $c_M=0$.  Indeed, if
$i$ is even, substitution in
$\widetilde u_{i-1}=(2i+y\widetilde u_i)^{-1}$ gives
$a_{i-1}=(2i+c_i)^{-1}$; if $i$ is odd, it gives
$c_{i-1}=a_i^{-1}$.  Hence
\begin{equation}
 c_{i-2}=c_i+2i.
 \label{eq:c-recurrence-app}
\end{equation}
For odd $\nu$,
\begin{equation}
 a_\nu^{-1}
 =2\sum_{i=\nu+1,\nu+3,\ldots,2\nu}i
 =\frac{(\nu+1)(3\nu+1)}{2},
\end{equation}
whereas for even $\nu$,
\begin{equation}
 c_\nu
 =2\sum_{i=\nu+2,\nu+4,\ldots,2\nu}i
 =\frac{\nu(3\nu+2)}{2}.
\end{equation}
Multiplication by $x$ proves
\eqref{eq:miller-odd-asymptotic} and
\eqref{eq:miller-even-asymptotic}.  Because the recurrence has finite depth,
the rational expansions preserve the displayed remainder orders.

To obtain the forward asymptotic, telescope
\begin{equation}
 \prod_{j=0}^{\nu-1}\widetilde u_j
 =\frac{b_\nu}{x^\nu b_0},
 \qquad
 \widetilde\Phi_{\nu,2\nu}
 =\log I_0+\sum_{j=0}^{\nu-1}\log\widetilde u_j.
 \label{eq:potential-product-app}
\end{equation}
For odd $\nu$, the range $j=0,\ldots,\nu-1$ contains
$(\nu+1)/2$ even indices.  Their terms in
\eqref{eq:parity-ansatz-app} each contribute
$-2\log x$ plus a constant; the remaining terms approach positive constants.
Combining this with
\begin{equation}
 \log I_0(x)=x-\frac12\log x+C_0+O(x^{-1})
\end{equation}
gives \eqref{eq:miller-forward-asymptotic}.  The finite rational expansions
and the standard Bessel expansion may be differentiated term by term, proving
\eqref{eq:miller-forward-derivative-asymptotic}.

For comparison,
\begin{equation}
 \phinu(x)=x-(\nu+1/2)\log x+C_\nu^\star+O(x^{-1}),
\end{equation}
and therefore
\begin{equation}
 E_{\nu,2\nu}(x)=-\log x+C_\nu^E+O(x^{-1}),
 \qquad
 E_{\nu,2\nu}'(x)=-x^{-1}+O(x^{-2}).
 \label{eq:E-asymptotic-app}
\end{equation}
For all sufficiently large $x$, \eqref{eq:miller-odd-asymptotic} implies
$\overline R_{\nu,2\nu}=1$.  Thus
\begin{align}
 D_{\mathrm{clip}}
 &=-\frac{\nu+3/2}{x}+O(x^{-2}),\\
 \overline R_{\nu,2\nu}-\Rnu
 &=\frac{\nu+1/2}{x}+O(x^{-2}),\\
 \widetilde\Phi_{\nu,2\nu}'-\Rnu
 &=-\frac1x+O(x^{-2}).
 \label{eq:clip-phase-app}
\end{align}
This completes the proof of Theorem~\ref{thm:miller-parity}.

\subsection{Finite-forward endpoint and defect identities}
\label{app:miller-score-proof}

By definition,
\begin{equation}
 \widetilde q_j-q_j
 =[\widetilde\Phi_{\nu,M}(r_j)-\phinu(r_j)]
 -[\widetilde\Phi_{\nu,M}(\kappa_j)-\phinu(\kappa_j)],
\end{equation}
which is \eqref{eq:miller-score-endpoint}.  The fundamental theorem of
calculus and \eqref{eq:defect-decomposition} give
\begin{align}
 \widetilde q_j-q_j
 &=\int_{\kappa_j}^{r_j}(\widetilde R_{\nu,M}-\Rnu)\,\mathrm dx
 +\int_{\kappa_j}^{r_j}D_{\mathrm{raw}}\,\mathrm dx,
 \label{eq:miller-score-raw-integral}\\
 &=\int_{\kappa_j}^{r_j}(\overline R_{\nu,M}-\Rnu)\,\mathrm dx
 +\int_{\kappa_j}^{r_j}D_{\mathrm{clip}}\,\mathrm dx.
 \label{eq:miller-score-clip-integral}
\end{align}
For odd $\nu$ with both endpoints in the high-concentration regime,
\begin{equation}
 \widetilde q_j-q_j
 =\log\frac{\kappa_j}{r_j}+O(\kappa_j^{-1}+r_j^{-1}).
 \label{eq:miller-score-asymptotic}
\end{equation}
Substitution of \eqref{eq:E-asymptotic-app} at both endpoints gives
\eqref{eq:miller-score-asymptotic}.

For completeness, the literal source-level stabilizer obeys
\begin{equation}
 -\nu\log(x+\epsilon)
 =-\nu\log x-\nu\log(1+\epsilon/x).
\end{equation}
Thus
\begin{equation}
 \widetilde\Phi_{\nu,M}^{\epsilon}(x)
 =\widetilde\Phi_{\nu,M}(x)-\nu\log(1+\epsilon/x),
 \label{eq:miller-epsilon-potential}
\end{equation}
and the literal endpoint error is
\begin{equation}
 \widetilde q_j^\epsilon-q_j
 =E_{\nu,M}(r_j)-E_{\nu,M}(\kappa_j)
 -\nu\!\left[\log(1+\epsilon/r_j)-\log(1+\epsilon/\kappa_j)\right].
 \label{eq:miller-score-epsilon}
\end{equation}
Taking an endpoint difference proves \eqref{eq:miller-score-epsilon};
differentiation gives the additional
$\nu\epsilon/[x(x+\epsilon)]$ term.  The extra term quantifies the source stabilizer separately from finite-start error.

\subsection{Proof of the AR uniform certificate}
\label{app:a2-proof}

We prove Theorem~\ref{thm:a2} using a Riccati residual and constant barriers.

Set $\varepsilon=\nu^{-1}$, $x=\nu z$, and
\begin{equation}
 H_\varepsilon(z)=\frac{I_{\nu+1}(\nu z)}{I_\nu(\nu z)}.
\end{equation}
The adjacent-order Bessel ratio satisfies
\begin{equation}
 \varepsilon H_\varepsilon'(z)
 =1-\frac{2+\varepsilon}{z}H_\varepsilon(z)
 -H_\varepsilon(z)^2.
 \label{eq:a2-riccati-app}
\end{equation}
Let $S=\sqrt{1+z^2}$ and define
\begin{equation}
\begin{split}
 r_0(z)&=\frac{z}{1+S},\\
 r_1(z)&=-\frac{z}{2S^2},\\
 r_2(z)&=\frac{z(4-z^2)}{8S^5},\\
 a_\varepsilon(z)&=r_0(z)+\varepsilon r_1(z)
 +\varepsilon^2r_2(z).
\end{split}
\label{eq:a2-candidate-app}
\end{equation}
Substituting $x=\nu z$ shows that
$a_\varepsilon(z)=\Atwo(x)$.

Define the residual obtained by inserting $a_\varepsilon$ into
\eqref{eq:a2-riccati-app}:
\begin{equation}
 \mathcal E
 =1-\frac{2+\varepsilon}{z}a_\varepsilon
 -a_\varepsilon^2-\varepsilon a_\varepsilon'.
\end{equation}
Direct simplification gives the exact identity
\begin{equation}
\begin{split}
 \mathcal E={}&-\varepsilon^3
 \frac{z^4-10z^2+4}{4(1+z^2)^{7/2}}\\
 &-\varepsilon^4
 \frac{z^2(z^2-4)^2}{64(1+z^2)^5}.
\end{split}
\label{eq:a2-exact-residual-app}
\end{equation}
For $u=z^2\ge0$,
\begin{equation}
 |u^2-10u+4|\le4(1+u)^2,
 \qquad
 u(u-4)^2\le16(1+u)^3.
\end{equation}
Consequently,
\begin{equation}
 |\mathcal E(z,\varepsilon)|
 \le\varepsilon^3(1+\varepsilon/4).
 \label{eq:a2-residual-bound-app}
\end{equation}

Let $e=H_\varepsilon-a_\varepsilon$.  Subtracting the two
Riccati equations yields
\begin{equation}
 \varepsilon e'=\mathcal E-Ke-e^2,
 \label{eq:a2-error-ode-app}
\end{equation}
where
\begin{equation}
\begin{split}
 K&=\frac{2+\varepsilon}{z}+2a_\varepsilon\\
 &=\frac{2S}{z}+\frac{\varepsilon}{zS^2}
 +2\varepsilon^2r_2(z).
\end{split}
\end{equation}
When $0<z\le2$, $r_2(z)\ge0$.  For $z\ge2$,
\begin{equation}
 -r_2(z)=\frac{z(z^2-4)}{8S^5}\le\frac1{40}.
\end{equation}
Therefore, for $0<\varepsilon\le9/10$,
\begin{equation}
 K\ge2-\frac{\varepsilon^2}{20}
 \ge\frac{3919}{2000}.
 \label{eq:a2-k-bound-app}
\end{equation}

Choose the constant barrier $b=\varepsilon^3$.  At $e=b$, using
\eqref{eq:a2-residual-bound-app} and \eqref{eq:a2-k-bound-app}, the
right-hand side of \eqref{eq:a2-error-ode-app} is strictly negative.  At
$e=-b$, it is bounded below by
\begin{equation}
 \varepsilon^3\left(
 -\frac{49}{40}+\frac{3919}{2000}-\frac{729}{1000}
 \right)
 =\frac{11}{2000}\varepsilon^3>0.
 \label{eq:a2-lower-margin-app}
\end{equation}
Thus the vector field points into the interval $[-b,b]$ on both boundaries.

It remains to initialize the barrier argument at the singular endpoint.  The
small-$z$ expansions are
\begin{equation}
 H_\varepsilon(z)=\frac{z}{2(1+\varepsilon)}+O(z^3)
\end{equation}
and
\begin{equation}
 a_\varepsilon(z)=\frac z2(1-\varepsilon+\varepsilon^2)+O(z^3).
\end{equation}
Hence
\begin{equation}
 e(z)=-\frac{\varepsilon^3}{2(1+\varepsilon)}z+O(z^3),
\end{equation}
so $e$ starts strictly between the two barriers.  If a first contact with the
upper barrier existed, it would require $e'\ge0$, contradicting the inward
direction above; a first contact with the lower barrier similarly requires
$e'\le0$ and gives a contradiction.  Therefore
$|e(z)|<\varepsilon^3$ for all $z>0$.  Both the exact ratio and AR extend
continuously to zero with value zero, yielding the non-strict result at
$z=0$.  Since $\varepsilon\le9/10$ is equivalent to $\nu\ge10/9$,
Theorem~\ref{thm:a2} follows. \qed

\begin{corollary}[Complexity--accuracy reversal]
\label{cor:complexity-certificate}
For integer feature dimension $p=2\nu+2$, the original start
$M=2\nu=p-2$ executes exactly $p-2$ dependent recurrence updates.  Under
Theorem~\ref{thm:a2}, our recurrence-free ratio approximation satisfies
\begin{equation}
 \sup_{x\ge0}|\Rnu(x)-\Atwo(x)|
 \le\nu^{-3}=\frac{8}{(p-2)^3}.
 \label{eq:complexity-certificate}
\end{equation}
Thus the original evaluator becomes sequentially deeper as $p$ grows, while
the FR/AR exact-arithmetic certificate tightens.  Both statements concern the evaluator as a function of feature dimension.
\end{corollary}

\subsection{Proof of the PATT-specific certificate}
\label{app:patt-proof}

The two-view augmentation loss satisfies
\begin{equation}
 |\widehat L_{\mathrm{ISAC}}-L_{\mathrm{ISAC}}|
 \le\overline\delta^{(2)}+\overline\delta^{(3)}.
 \label{eq:patt-loss-bound}
\end{equation}
Under Theorem~\ref{thm:patt}, put $\eta=(\tau\nu^3)^{-1}$.  For each sample and
view, the adjusted Softmax distributions satisfy
\begin{equation}
 \TV(P_i^{(v)},\widehat P_i^{(v)})
 \le\tanh(\delta_i^{(v)}/2),
 \label{eq:patt-tv}
\end{equation}
an exact adjusted top-two margin greater than $2\delta_i^{(v)}$ preserves the
per-view score prediction, and
\begin{equation}
\left\|\nabla_{f_i^{(v)}}\widehat L_{\mathrm{ISAC}}
-\nabla_{f_i^{(v)}}L_{\mathrm{ISAC}}\right\|
\le\frac{1}{2B}\left[
\frac{2}{\tau}\tanh\left(\frac{\delta_i^{(v)}}2\right)+2\eta
\right].
\label{eq:patt-grad-bound}
\end{equation}

\begin{corollary}[Head, medium, and tail groups]
\label{cor:groups}
For a nonempty sample subset $G$, let
$L_G^{(v)}=|G|^{-1}\sum_{i\in G}\ell_i^{(v)}$ and
$L_G=(L_G^{(2)}+L_G^{(3)})/2$.  Then
\begin{align}
 |\widehat L_G^{(v)}-L_G^{(v)}|
 &\le\frac{2}{|G|}\sum_{i\in G}\delta_i^{(v)},
 \label{eq:group-loss-bound}\\
 |\widehat L_G-L_G|
 &\le\frac{1}{|G|}\sum_{i\in G}(\delta_i^{(2)}+\delta_i^{(3)}).
 \label{eq:group-two-view-bound}
\end{align}
For
$G_i^{(v)}=2\tau^{-1}\tanh(\delta_i^{(v)}/2)+2\eta$,
\begin{equation}
 \left\|\nabla_{f_i^{(v)}}\widehat L_G
 -\nabla_{f_i^{(v)}}L_G\right\|
 \le\frac{G_i^{(v)}}{2|G|},\qquad i\in G.
 \label{eq:group-gradient-bound}
\end{equation}
\end{corollary}

Since $\phinu'=\Rnu$ and $\Ftwo'=\Atwo$,
\begin{equation}
 q_j(f)-\widehat q_j(f)
 =\int_{\kappa_j}^{r_j(f)}[\Rnu(x)-\Atwo(x)]\,\mathrm dx.
\end{equation}
Theorem~\ref{thm:a2} and the reverse triangle inequality yield
\begin{equation}
 |q_j(f)-\widehat q_j(f)|
 \le\frac{|r_j(f)-\kappa_j|}{\nu^3}
 \le\frac{\norm f}{\tau\nu^3}.
 \label{eq:score-proof-app}
\end{equation}

For one sample and view, let $s_j=q_j+b_j$ and
$\widehat s_j=s_j+e_j$, where $|e_j|\le\delta$.  Monotonicity of log-sum-exp
under coordinatewise bounds gives
\begin{equation}
 \min_j e_j\le
 \log\sum_j e^{s_j+e_j}-\log\sum_j e^{s_j}
 \le\max_j e_j.
\end{equation}
After subtracting the target-score perturbation,
\begin{equation}
 |\widehat\ell_y-\ell_y|
 \le\max_j e_j-\min_j e_j\le2\delta.
 \label{eq:ce-proof-app}
\end{equation}
The biases $b_j$ are arbitrary and the argument introduces no sum over
classes.  Averaging \eqref{eq:ce-proof-app} over the batch and the two views
proves \eqref{eq:patt-loss-bound}.  Because TLA and OE are unchanged, their
difference is zero and \eqref{eq:patt-total-bound} follows.

Let $P=\operatorname{softmax}(s)$ and
$\widehat P=\operatorname{softmax}(s+e)$.  Reweighting $P_j$ by factors in
$[e^{-\delta},e^\delta]$ gives
$\TV(P,\widehat P)\le\tanh(\delta/2)$.  A pairwise adjusted-score margin
changes by at most $2\delta$, which proves the prediction statement.

For $v_j=\kappa_j\mu_j+f/\tau$ and $r_j=\norm{v_j}>0$,
\begin{equation}
 \nabla_fq_j=\frac1\tau\Rnu(r_j)\frac{v_j}{r_j},
 \qquad
 \nabla_f\widehat q_j=\frac1\tau\Atwo(r_j)\frac{v_j}{r_j}.
\end{equation}
Theorem~\ref{thm:a2} therefore gives
$\norm{\nabla_fq_j-\nabla_f\widehat q_j}\le\eta$; both gradients extend
continuously at $r_j=0$.  Writing a one-sample cross-entropy gradient as the
Softmax-weighted score-gradient average minus the target gradient, adding and
subtracting the exact gradients under $\widehat P$, and using
$\norm{\widehat P-P}_1\le2\tanh(\delta/2)$ yields
\begin{equation}
 \norm{\nabla_f\widehat\ell_y-\nabla_f\ell_y}
 \le\frac2\tau\tanh(\delta/2)+2\eta.
\end{equation}
The batch mean contributes $1/B$ and the two-view average contributes $1/2$,
proving \eqref{eq:patt-grad-bound}.  For a nonempty group $G$,
\eqref{eq:ce-proof-app} averaged within one view gives
\eqref{eq:group-loss-bound}; averaging its two view-specific bounds gives
\eqref{eq:group-two-view-bound}.  The single-sample, single-view gradient bound
is $G_i^{(v)}$.  In $L_G$, the feature $f_i^{(v)}$ appears with coefficient
$1/(2|G|)$, which proves \eqref{eq:group-gradient-bound}.

\section{Numerical validation and mechanism analysis}
\label{app:numerical-validation}
\subsection{Frozen-state mechanism validation}
\label{app:mechanism-protocol}

Frozen-state diagnostics hold the PATT features, augmented views, and class state fixed, isolating the backward rule from state updates and geometry changes.

\paragraph{Concentration and depth scans.}
We vary feature dimension, scaled concentration, and recurrence depth as in Figure~\ref{fig:proco-conditions}, and test adjacent odd and even orders. Independent high-precision and finite-difference checks reproduce the predicted parity and finite-forward derivative. Increasing recurrence depth can improve raw-ratio fidelity while worsening clipped coherence, consistent with their distinct definitions. The scan covers $x/\nu\in\{0.25,0.5,1,2,4,8,16,32,64,128,256,512,1024\}$ and $M/\nu\in\{2,3,4,6,8\}$ at $p\in\{128,256,512,1024\}$. The marker locates the median query radius from one captured PATT state ($p=512$, $M/\nu=2$); its query range is $r/\nu\in[392.35,393.11]$.

\paragraph{Frozen-state backward comparison.}
Original recurrence uses the saved raw ratio and unit-clipped backward; consistent recurrence differentiates the identical finite-start forward.  Table~\ref{tab:backward-counterfactual} shows exact score/loss
agreement but a nonzero radial defect and aggregated feature-gradient
difference in both views.  Thus changing backward alone exposes the local
field mismatch without a forward confound.

\begin{table*}[!htbp]
\centering
\caption{\textbf{Changing only the backward rule reveals a PATT coherence defect.}
Original and consistent recurrence use the same frozen CIFAR100-LT state and two augmented views (seed 3407). Score and loss columns report absolute differences; gradient columns compare the complete adjusted cross-entropy feature gradients. $D_{\rm clip}=\widetilde\Phi'-g=-d$ is the signed radial residual.}
\label{tab:backward-counterfactual}
\small
\setlength{\tabcolsep}{6pt}
\begin{tabular}{lrrrrr}
\toprule
View & Max score $|\Delta|$ & Loss $|\Delta|$ & Max $|D_{\rm clip}|$
& Gradient relative $\ell_2$ & Gradient max $|\Delta|$ \\
\midrule
2 & $0$ & $0$ & $2.55\times10^{-3}$
& $2.56\times10^{-3}$ & $1.40\times10^{-5}$ \\
3 & $0$ & $0$ & $2.55\times10^{-3}$
& $2.56\times10^{-3}$ & $1.35\times10^{-5}$ \\
\bottomrule
\end{tabular}
\end{table*}

\paragraph{Odd/even dimensional behavior.}
Table~\ref{tab:parity-counterfactual} confirms linear growth for odd orders and inverse decay for even orders, in agreement with Equations~\eqref{eq:miller-odd-asymptotic}--\eqref{eq:miller-even-asymptotic}.

\begin{table*}[!htbp]
\centering
\caption{\textbf{Finite-recurrence asymptotics follow the predicted parity.}
At $x=10^9$ and $M=2\nu$, odd and even orders are compared with their analytic limits using 100-digit arithmetic. The scaled quantity is $\widetilde R/x$ for odd $\nu$ and $x\widetilde R$ for even $\nu$; sign refers to $\widetilde R-\Rnu$.}
\label{tab:parity-counterfactual}
\small
\setlength{\tabcolsep}{3pt}
\begin{tabular*}{0.90\linewidth}{@{\extracolsep{\fill}}rrrrlrrr@{}}
\toprule
$\nu$ & $p$ & Raw $\widetilde R$ & Sign & Phase & Scaled & Limit & Rel. error \\
\midrule
63  & 128 & $1.645\times10^{5}$  & $+$ & linear  & $1.645\times10^{-4}$ & $1.645\times10^{-4}$ & $1.2\times10^{-11}$ \\
64  & 130 & $6.208\times10^{-6}$  & $-$ & inverse & $6.208\times10^{3}$  & $6.208\times10^{3}$  & $1.3\times10^{-11}$ \\
255 & 512 & $1.020\times10^{4}$  & $+$ & linear  & $1.020\times10^{-5}$ & $1.020\times10^{-5}$ & $3.2\times10^{-9}$ \\
256 & 514 & $9.856\times10^{-5}$ & $-$ & inverse & $9.856\times10^{4}$ & $9.856\times10^{4}$  & $3.3\times10^{-9}$ \\
\bottomrule
\end{tabular*}
\end{table*}

\paragraph{Robustness across precision and geometry.}
We cross FP32/FP64 geometry with expanded/factorized evaluation on the same frozen view. Within every cell, original and consistent recurrence retain identical scores and losses, while their feature gradients differ. The mismatch therefore persists across the precision and geometry choices.

\subsection{Fidelity benchmarks and numerical references}
\label{app:detailed-tables}
\label{app:non-timing}
Table~\ref{tab:captured-audit} separates numerical score and derivative discrepancies at captured learning states.
\begin{table}[H]
\centering
\caption{\textbf{Score fidelity and backward coherence at captured PATT states.}
Maximum absolute discrepancies over six CIFAR100-LT views. Score and radial fidelity use the exact Bessel reference; coherence compares the supplied backward with the derivative of its own finite forward.}
\label{tab:captured-audit}
\small
\begin{tabular}{@{}llr@{}}
\toprule
Comparison & Reference & Maximum discrepancy \\
\midrule
Finite-potential score & Exact score & $3.32\times10^{-3}$ \\
AR/FR same-state score & Exact score & $1.17\times10^{-9}$ \\
Clipped backward radial factor & Exact Bessel ratio & $2.55\times10^{-3}$ \\
Forward--backward coherence defect & Finite-forward derivative & $2.56\times10^{-3}$ \\
\bottomrule
\end{tabular}
\end{table}

The three tables address complementary levels of fidelity. Table~\ref{tab:captured-audit} evaluates captured learning states, Table~\ref{tab:dimension-main} isolates the adjacent-order ratio over a controlled concentration grid, and Table~\ref{tab:endpoint-scaling} evaluates the endpoint potential differences that enter the learning score. Together they separate state-dependent implementation effects from the intrinsic approximation error of AR/FR.

\paragraph{Numerical references and implementation.}
The dimension study uses seven dimensions and 61 common scaled concentrations
per dimension. Reference Bessel values are evaluated with mpmath at 60 decimal
digits. Endpoint scores use $\Phi_\nu(x+1)-\Phi_\nu(x)$, so the score certificate
is $\nu^{-3}$. The production AR/FR implementation uses cancellation-resistant
endpoint differences in FP64. Stable log-Miller uses the original start
$2\nu$ and normalization at order zero; logarithmic arithmetic isolates the
finite-start effect from intermediate overflow. Its ratio column uses the
clipped adjacent-order ratio.
For CUSF, adjacent-order values define the ratio. CUSF and log-Miller errors use FP64 outputs compared with the common reference.
The AR/FR approximation errors are evaluated with 60-digit arithmetic.

\begin{table}[t]
\centering
\caption{\textbf{Ratio fidelity across feature dimensions.}
Maximum absolute errors use 61 log-spaced concentrations $x/\nu\in[10^{-3},10^3]$ per dimension and a 60-digit Bessel reference. CUSF forms a ratio from adjacent log-Bessel values in CPU FP64; AR/FR is evaluated in 60-digit arithmetic. The certificate $\nu^{-3}$ covers the full nonnegative axis.}
\label{tab:dimension-main}
\small
\begin{tabular}{@{}rrrr@{}}
\toprule
Dimension & CUSF & AR/FR & AR/FR certificate \\
\midrule
64 & $3.37\times10^{-12}$ & $2.74\times10^{-6}$ & $3.36\times10^{-5}$ \\
128 & $2.38\times10^{-11}$ & $3.29\times10^{-7}$ & $4.00\times10^{-6}$ \\
256 & $2.29\times10^{-11}$ & $4.03\times10^{-8}$ & $4.88\times10^{-7}$ \\
512 & $1.85\times10^{-11}$ & $4.99\times10^{-9}$ & $6.03\times10^{-8}$ \\
1024 & $2.45\times10^{-11}$ & $6.21\times10^{-10}$ & $7.49\times10^{-9}$ \\
2048 & $8.87\times10^{-11}$ & $7.74\times10^{-11}$ & $9.34\times10^{-10}$ \\
4096 & $5.76\times10^{-10}$ & $9.66\times10^{-12}$ & $1.17\times10^{-10}$ \\
\bottomrule
\end{tabular}
\end{table}

\begin{table}[t]
\centering
\caption{\textbf{Fidelity of unit-length potential differences.}
Maximum absolute errors in $\Phi_\nu(x+1)-\Phi_\nu(x)$ use the common 61-point concentration grid at each dimension. Log-Miller retains the finite start; CUSF subtracts separately evaluated potentials; AR/FR uses its paired analytic potential. All errors use the high-precision Bessel reference.}
\label{tab:endpoint-scaling}
\small
\begin{tabular}{@{}rrrr@{}}
\toprule
Dimension & Log-Miller & CUSF & AR/FR \\
\midrule
64 & $3.36\times10^{-4}$ & $1.89\times10^{-12}$ & $2.76\times10^{-6}$ \\
128 & $8.29\times10^{-5}$ & $8.47\times10^{-12}$ & $3.30\times10^{-7}$ \\
256 & $2.06\times10^{-5}$ & $2.31\times10^{-11}$ & $4.04\times10^{-8}$ \\
512 & $5.13\times10^{-6}$ & $9.72\times10^{-12}$ & $5.00\times10^{-9}$ \\
1024 & $1.28\times10^{-6}$ & $3.21\times10^{-11}$ & $6.21\times10^{-10}$ \\
2048 & $2.11\times10^{-7}$ & $9.67\times10^{-11}$ & $7.74\times10^{-11}$ \\
4096 & $7.25\times10^{-9}$ & $1.73\times10^{-10}$ & $9.67\times10^{-12}$ \\
\bottomrule
\end{tabular}
\end{table}

\FloatBarrier
\subsection{Learning-level stress tests}
\paragraph{Class count and prior imbalance.}
We generate eight normalized 128-dimensional queries and a pool of 2000
class directions with seed 3407. Concentrations span 10 to 10,000; smaller
class sets are prefixes of the same randomly ordered pool. Each class set
uses log-linear priors with largest-to-smallest ratios 1, 100, and 1000.
All 15 configurations satisfy the score bound $4.00\times10^{-5}$ and the
mean cross-entropy bound $8.00\times10^{-5}$. 

\begin{figure}[!htbp]\centering
\includegraphics[width=\linewidth]{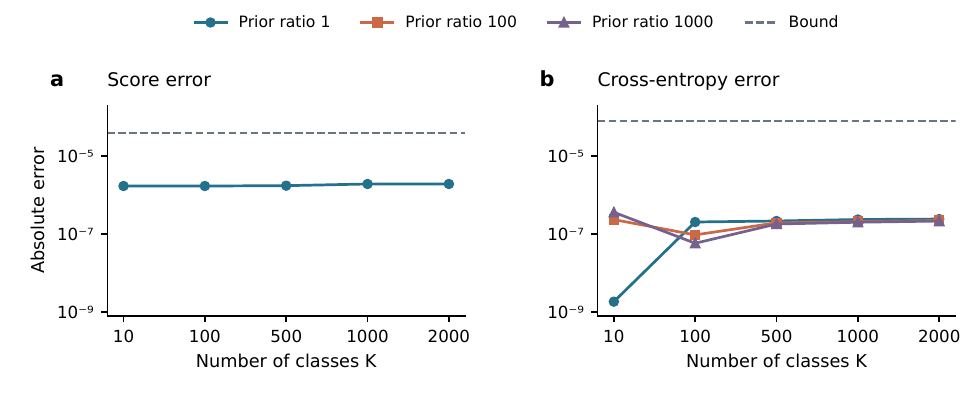}
\caption{\textbf{Score and loss errors remain below their certificates as class count varies.}
The 15 fixed-state configurations combine five class counts with prior ratios 1, 100, and 1000. (a) The three maximum score-error curves coincide and are drawn once. (b) Mean cross-entropy error is shown separately for each prior ratio. Dashed lines denote the respective theoretical bounds. Class counts are equally spaced discrete settings.}
\label{tab:class-count-control}
\label{fig:class-count-control}
\end{figure}

\Needspace{0.66\textheight}
\subsection{Measured PATT update fields}
\label{app:experiment-details}
\begin{figure}[H]\centering
\includegraphics[width=0.90\linewidth]{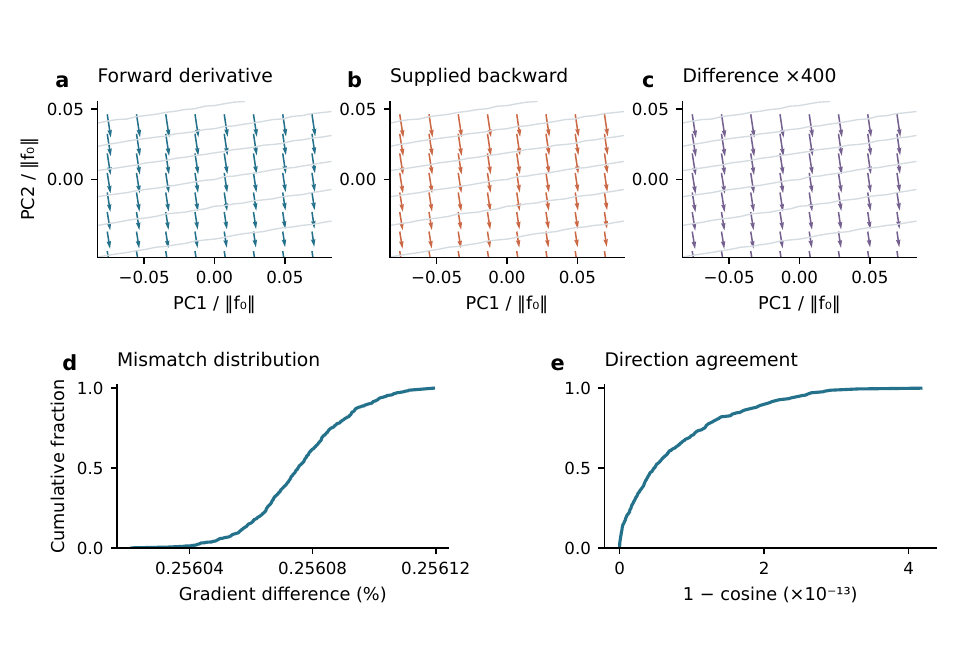}
\caption{\textbf{A frozen PATT state exhibits mainly gradient-magnitude distortion.}
The first CIFAR-10 query is varied over a $25\times25$ slice along two principal feature directions, with class statistics fixed. Scores and losses agree bitwise at every location.
(a--b) Negative gradients from the finite forward and supplied backward, using common contours and arrow scales.
(c) Their difference magnified $400\times$; each field displays 64 interior arrows.
(d--e) Empirical cumulative distributions over all 625 projected-gradient pairs. Relative mismatch has median $0.2561\%$; $1-\cos$ resolves the small directional difference.}
\label{fig:patt-measured-field}
\end{figure}

Figure~\ref{fig:patt-measured-field} separates the supplied update from the derivative of the same finite forward on a real feature slice.

\FloatBarrier
\subsection{Classwise geometry and full-update behavior}
\label{app:mechanism-interventions}
\begin{figure}[!htbp]\centering
\includegraphics[width=\linewidth]{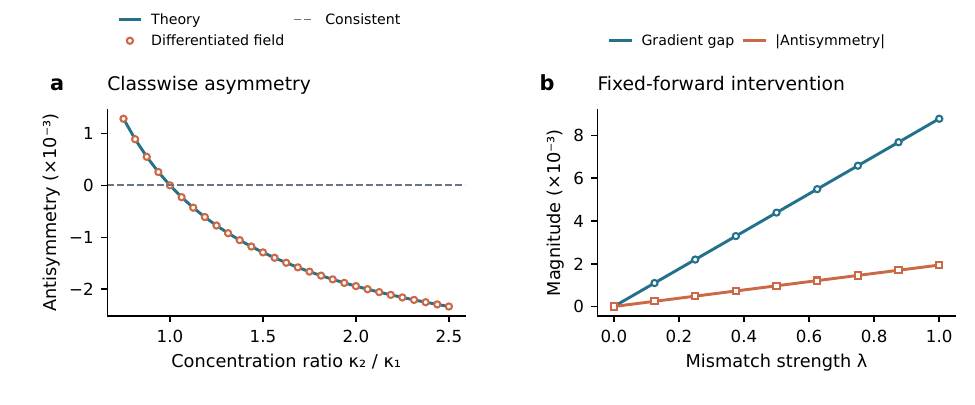}
\caption{\textbf{Classwise mismatch controls the local Jacobian obstruction.}
(a) Signed antisymmetry versus concentration ratio at 29 constructed states: lines show the analytic prediction and circles the differentiated field; the consistent field remains zero.
(b) Gradient-gap norm and absolute antisymmetry across nine mismatch strengths, with predictions shown as lines and computed values as markers. All interventions retain forward loss $\log 2$.}
\label{fig:mechanism-controls}
\end{figure}

We vary the classwise concentration ratio in the two-class construction of Appendix~\ref{app:curl-proof}. The resulting Jacobian antisymmetry follows the analytic prediction and vanishes at equal concentrations. A complementary fixed-forward intervention varies the mismatch strength and yields the predicted linear response in both gradient gap and antisymmetry (Figure~\ref{fig:mechanism-controls}).

\paragraph{Clipping and the high-concentration boundary.}
For each $\nu\in\{63,255\}$ we evaluate concentrations
$x/\nu\in[0.25,10^5]$ at $M=2\nu$. The clipped defect eventually decreases
as $(\nu+3/2)/x$, despite divergence of the raw ratio. This distinguishes
raw ratio fidelity from clipped coherence. The saved PATT query interval
$x/\nu\in[392.35,393.11]$ lies inside the expanded range.

\begin{figure}[!htbp]\centering
\includegraphics[width=\linewidth]{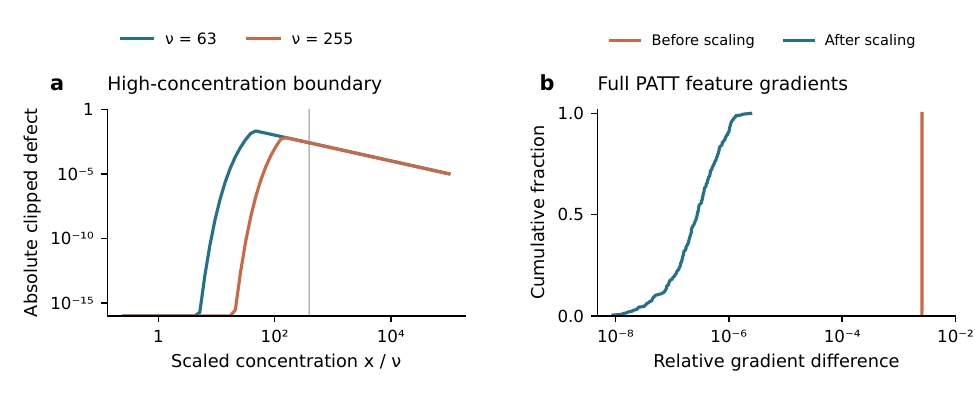}
\caption{\textbf{Concentration and gradient scaling delimit the coherence defect.}
(a) Absolute clipped defect over 130 conditions; dashed segments show the asymptote $(\nu+3/2)/x$, and the grey interval marks saved PATT query radii. Values below $10^{-16}$ share a display floor.
(b) Cumulative distributions for 256 frozen PATT queries in the full feature space with FP64 geometry. Relative differences use the consistent gradient norm before scaling and the supplied gradient norm after optimal per-query scaling.}
\label{fig:regime-scaling}
\end{figure}

\paragraph{Full feature gradients and removal of overall scaling.}
We reconstruct all 256 saved PATT queries in FP64 at the same fixed class
bank, using the complete 512-dimensional features, original labels, priors,
and saved normalizer offsets. For consistent gradient $a_i$ and supplied
gradient $b_i$, set $c_i^*=a_i^\top b_i/\|a_i\|^2$ and measure
$\|b_i-c_i^*a_i\|/\|b_i\|$.
At this frozen state, the discrepancy is dominated by gradient magnitude: optimal per-query scalar alignment leaves a much smaller residual (Figure~\ref{fig:regime-scaling}).

\paragraph{Full-parameter update.}
A one-step replay from a shared model, optimizer, and class state preserves identical forward scores and losses but yields different total parameter gradients and optimizer updates. A global rescaling of the consistent gradient does not eliminate the residual, showing that the effect is not explained by a single multiplicative factor.

\FloatBarrier
\section{Structure-aware execution and efficiency}
\subsection{Paired evaluation in the host learning system}
\label{sec:system}
AR/FR replaces the repeated vMF score and its radial derivative. The host class-state estimand, detached-statistics semantics, view order, priors, auxiliary losses, encoder, and optimizer retain their original definitions.

\paragraph{Matched training realizations.}
The three-arm CIFAR experiment shares FP32 expanded geometry and backbone, FP64 scalar evaluation, and disabled autocast. Original and consistent recurrence use the same finite forward. AR/FR replaces the score pair; all branches retain the common class-bank update.

\subsubsection{Finite-sample state identities}
\label{app:state-proof}

\begin{lemma}[Finite-sample resultant decomposition]
\label{lem:finite-state}
For realized $z_1,\ldots,z_n\in\mathbb R^p$, $n\ge1$, write
$R_n=\lVert n^{-1}\sum_a z_a\rVert$.  Then
\begin{equation}
 R_n^2=\frac1{n^2}\sum_a\norm{z_a}^2
 +\frac1{n^2}\sum_{a\ne b}z_a^\top z_b.
 \label{eq:resultant-second-order}
\end{equation}
If the vectors are i.i.d. with mean $m$ and covariance $\Sigma$, then
\begin{equation}
 \mathbb ER_n^2=\norm m^2+\frac{\operatorname{tr}\Sigma}{n};
 \label{eq:resultant-expectation}
\end{equation}
for unit-norm features this is $n^{-1}+(1-n^{-1})\norm m^2$.
\end{lemma}

\begin{proposition}[Exact sum--count state reduction]
\label{prop:state-reduction}
For current count $N\ge0$ and mean $a$, and batch count $m\ge0$ and sum $s$,
the original update satisfies
\begin{equation}
 a^+=\frac{N}{N+m}a+\frac{s}{N+m}=\frac{Na+s}{N+m},
 \qquad N^+=N+m,
 \label{eq:sum-count-update}
\end{equation}
whenever $N+m>0$; if $m=0$, the state is unchanged.
\end{proposition}

Expanding the squared norm gives
\begin{equation}
 \left\|\frac1n\sum_{a=1}^n z_a\right\|^2
 =\frac1{n^2}\sum_{a=1}^n\sum_{b=1}^n z_a^\top z_b.
\end{equation}
Separating diagonal and off-diagonal terms proves
\eqref{eq:resultant-second-order}.  If $z_1,\ldots,z_n$ are i.i.d. with mean
$m$ and covariance $\Sigma$, then the sample mean has mean $m$ and covariance
$\Sigma/n$.  The identity
$\mathbb E\|X\|^2=\|\mathbb EX\|^2+\operatorname{tr}\operatorname{Cov}(X)$
therefore proves \eqref{eq:resultant-expectation}.  For unit-norm vectors,
$\operatorname{tr}\Sigma=\mathbb E\|z\|^2-\|m\|^2=1-\|m\|^2$.
No independence statement is used for the realized decomposition.

For Proposition~\ref{prop:state-reduction}, the original update first forms the
batch average $s/m$ when $m>0$ and uses weight $m/(N+m)$.  Direct substitution
gives
\begin{equation}
 \left(1-\frac{m}{N+m}\right)a
 +\frac{m}{N+m}\frac{s}{m}
 =\frac{Na+s}{N+m}.
\end{equation}
When $m=0$, the source sets the weight to zero and leaves the state unchanged.
When $N=0<m$, the formula reduces to $s/m$.  When $N=m=0$, the data-derived
mean is undefined and the source keeps its initializer.  Thus class sums and
counts are sufficient to reproduce this deterministic state transition.

\subsection{Stable potential differences and factorized execution}
\label{app:system}

The original per-view state-update order and stop-gradient semantics are preserved.

\paragraph{Direct sufficient-statistic accumulation.}
Class sums and counts are accumulated directly rather than through a dense $B\times K\times p$ assignment expansion. This preserves the sufficient statistics in Proposition~\ref{prop:state-reduction} while reducing intermediate storage.

\paragraph{Factorized vMF geometry.}
Feature norms and a $B\times K$ projection matrix determine the score geometry without materializing classwise resultant vectors. Stable endpoint differences avoid cancellation between large potentials; the expressions below are evaluated in double precision.

\paragraph{Fixed-depth compiled execution.}
The paired evaluator has a fixed operation structure suitable for compilation and GPU graph replay. This reduces launch overhead in addition to the storage savings from statistics and geometry; compilation is excluded from steady-state timing.

The analytic potential is
\begin{equation}
\begin{split}
 \Ftwo(x)={}&s(x)-\nu\log(\nu+s(x))-\frac12\log s(x)\\
 &+\frac{3x^2-2\nu^2}{24s(x)^3},\qquad
 s(x)=\sqrt{\nu^2+x^2},
\end{split}
\label{eq:f2}
\end{equation}
The stable endpoint increment is
\begin{equation}
 d_j=r_j^2-\kappa_j^2
 =\frac{2\kappa_j\mu_j^\top f}{\tau}+\frac{\norm f^2}{\tau^2},
 \quad
 \Delta_j=\frac{d_j}{s(r_j)+s(\kappa_j)},
 \label{eq:stable-delta}
\end{equation}

With $s_1=s(r_j)$ and $s_0=s(\kappa_j)$, the cancellation-resistant score is
\begin{equation}
\begin{split}
\widehat q_j={}&\Delta_j
-\nu\log\left(1+\frac{\Delta_j}{\nu+s_0}\right)
-\frac12\log\left(1+\frac{\Delta_j}{s_0}\right)\\
&-\frac{\Delta_j}{8s_1s_0}
+\frac{5\nu^2\Delta_j(s_1^2+s_1s_0+s_0^2)}
{24s_1^3s_0^3}.
\end{split}
\label{eq:stable-f2}
\end{equation}

\subsection{Unified H100 local runtime and memory protocol}
\label{app:h100-local}
Figure~\ref{fig:efficiency-compact} uses a fixed state and synthetic inputs with $B=32$, $K=1000$, $p=1024$, and seed 3407. The local boundary includes two-view class-state updates, score geometry, cross-entropy, and backward to features; it excludes the backbone and data loading. Each method uses seven timed repetitions after three warmups, with state restoration outside measurement. Compilation is excluded from steady-state timing. Structure-aware AR/FR uses compiled statistics and score geometry with explicit CUDA Graph replay.

Memory is the peak allocated increment within the local execution boundary above prepared inputs and warmed modules, including persistent allocations created by explicit graph capture and the replay peak. This metric measures additional local storage rather than full-process VRAM. The five forward--backward measurements are summarized in Table~\ref{tab:h100-local}.

\begin{table}[!htbp]\centering\small
\caption{\textbf{Local H100 forward--backward costs.} Median runtime and incremental peak allocation under the protocol in Appendix~\ref{app:h100-local}.}
\label{tab:h100-local}
\begin{tabular}{@{}lrr@{}}\toprule
Numerical realization & Runtime (ms) & Peak increment (MiB)\\\midrule
Original recurrence & 266.40 & 504.89\\
Consistent recurrence & 412.80 & 1755.42\\
Log-domain Miller & 120.28 & 504.89\\
AR/FR & 4.90 & 508.12\\
Structure-aware AR/FR & 0.206 & 32.00\\\bottomrule
\end{tabular}
\end{table}

\paragraph{Attribution of the combined gain.}
The dense-layout comparison isolates evaluator replacement: runtime falls from 266.40 to 4.90\,ms ($54.4\times$), while the peak increment remains comparable (504.89 versus 508.12\,MiB). Structure-aware execution then reduces runtime from 4.90 to 0.206\,ms ($23.8\times$) and the peak increment to 32.00\,MiB. Relative to original recurrence, the combined reductions are $1293\times$ in runtime and $15.8\times$ in memory. The second comparison measures statistics, geometry, and compiled execution jointly; it does not isolate their individual contributions.

\clearpage
\section{Extended task results}
\label{app:extended-tasks}
\subsection{Complete task metrics}
\label{app:aggregate-results}
Table~\ref{tab:task-complete} includes all three task metrics, complementing the compact main-text table.
\begin{table}[!htbp]\centering\small
\caption{\textbf{Complete classification and OOD detection results.} Metrics are percentages; CIFAR results report mean $\pm$ standard deviation over three runs. Bold and underline indicate the best and second-best values within each setting.}
\label{tab:task-complete}
\setlength{\tabcolsep}{4pt}
\begin{tabular}{@{}ll l ccc@{}}\toprule
Dataset & IF & Metric & Original recurrence & Consistent recurrence & AR/FR\\\midrule
CIFAR-10-LT & 10 & Acc. $\uparrow$ & $92.04\pm 0.17$ & $\underline{92.12}\pm 0.18$ & $\mathbf{92.16}\pm 0.19$ \\
 &  & AUROC $\uparrow$ & $\underline{93.07}\pm 0.13$ & $92.30\pm 0.27$ & $\mathbf{93.43}\pm 0.46$ \\
 &  & FPR95 $\downarrow$ & $\underline{26.20}\pm 1.19$ & $29.52\pm 1.06$ & $\mathbf{25.52}\pm 1.01$ \\
\midrule
CIFAR-10-LT & 50 & Acc. $\uparrow$ & $87.30\pm 0.29$ & $\mathbf{87.58}\pm 0.35$ & $\underline{87.42}\pm 0.15$ \\
 &  & AUROC $\uparrow$ & $90.18\pm 2.46$ & $\underline{90.95}\pm 1.57$ & $\mathbf{91.02}\pm 1.13$ \\
 &  & FPR95 $\downarrow$ & $34.51\pm 6.96$ & $\underline{32.84}\pm 1.94$ & $\mathbf{32.23}\pm 2.89$ \\
\midrule
CIFAR-10-LT & 100 & Acc. $\uparrow$ & $83.82\pm 0.21$ & $\underline{84.67}\pm 0.41$ & $\mathbf{84.73}\pm 0.31$ \\
 &  & AUROC $\uparrow$ & $\mathbf{90.66}\pm 0.13$ & $\underline{89.97}\pm 0.39$ & $89.95\pm 0.34$ \\
 &  & FPR95 $\downarrow$ & $\mathbf{33.35}\pm 0.50$ & $\underline{34.33}\pm 0.71$ & $35.15\pm 0.79$ \\
\midrule
CIFAR-100-LT & 10 & Acc. $\uparrow$ & $\underline{62.37}\pm 2.43$ & $62.12\pm 1.84$ & $\mathbf{62.56}\pm 2.03$ \\
 &  & AUROC $\uparrow$ & $\underline{78.18}\pm 2.09$ & $78.08\pm 1.23$ & $\mathbf{78.44}\pm 1.53$ \\
 &  & FPR95 $\downarrow$ & $57.74\pm 3.43$ & $\underline{57.70}\pm 1.46$ & $\mathbf{57.14}\pm 2.35$ \\
\midrule
CIFAR-100-LT & 50 & Acc. $\uparrow$ & $\underline{53.26}\pm 0.19$ & $52.87\pm 0.31$ & $\mathbf{53.74}\pm 0.31$ \\
 &  & AUROC $\uparrow$ & $75.37\pm 0.68$ & $\underline{75.54}\pm 0.60$ & $\mathbf{76.14}\pm 0.24$ \\
 &  & FPR95 $\downarrow$ & $\underline{62.77}\pm 1.00$ & $64.12\pm 0.96$ & $\mathbf{62.59}\pm 0.96$ \\
\midrule
CIFAR-100-LT & 100 & Acc. $\uparrow$ & $\mathbf{50.33}\pm 0.16$ & $49.48\pm 0.26$ & $\underline{49.66}\pm 0.16$ \\
 &  & AUROC $\uparrow$ & $\mathbf{75.78}\pm 0.33$ & $\underline{75.40}\pm 0.54$ & $75.21\pm 0.65$ \\
 &  & FPR95 $\downarrow$ & $\mathbf{63.56}\pm 0.25$ & $\underline{64.58}\pm 0.62$ & $64.91\pm 0.65$ \\
\midrule
ImageNet-LT & --- & Acc. $\uparrow$ & $53.88$ & $\underline{54.11}$ & $\mathbf{54.34}$ \\
 &  & AUROC $\uparrow$ & $76.19$ & $\underline{76.87}$ & $\mathbf{77.50}$ \\
 &  & FPR95 $\downarrow$ & $79.66$ & $\underline{78.74}$ & $\mathbf{76.68}$ \\
\bottomrule
\end{tabular}
\end{table}

\FloatBarrier

\end{document}